\documentclass[sigconf, nonacm]{acmart}
\newcommand{\eat}[1]{}
\usepackage{alltt}
\usepackage{latexsym}
\usepackage{amsfonts}
\usepackage{amsmath}
\usepackage{color}
\usepackage{colortbl}
\usepackage{epsfig}
\usepackage{xspace}
\usepackage{graphicx}
\usepackage{subfigure}
\usepackage{enumerate}
\usepackage{comment}
\usepackage{mathtools}

\usepackage{epsfig}
\usepackage{multirow}
\usepackage{hhline}
\usepackage{url}
\usepackage{framed}
\usepackage{xcolor}
\usepackage{geometry}

\newcommand{\system}{{RPT}\xspace}

\newcommand{\at}[1]{\textbf{\texttt{#1}}\xspace}
\newcommand{\stab}{\vspace{1.2ex}\noindent}
\newcommand{\sstab}{\rule{0pt}{8pt}\\[-2.2ex]}

\newcommand{\ra}{\rightarrow}

\newcommand{\bi}{\begin{itemize}}
\newcommand{\ei}{\end{itemize}}
\newcommand{\be}{\begin{enumerate}}
\newcommand{\ee}{\end{enumerate}}

\newcommand{\stitle}[1]{\vspace{1ex}\noindent{\bf #1}}
\newcommand{\etitle}[1]{\vspace{0.8ex}\noindent{\underline{\em #1}}}
\newcommand{\ie}{\emph{i.e.,}\xspace}
\newcommand{\eg}{\emph{e.g.,}\xspace}
\newcommand{\wrt}{\emph{w.r.t.}\xspace}

\newcommand{\M}{[{\bf M}]}
\newcommand{\AT}{[{\bf A}]}
\newcommand{\VT}{[{\bf V}]}

\newcommand\vldbdoi{10.14778/3457390.3457391}
\newcommand\vldbpages{XXX-XXX}
\newcommand\vldbvolume{14}
\newcommand\vldbissue{8}
\newcommand\vldbyear{2021}
\newcommand\vldbauthors{\authors}
\newcommand\vldbtitle{\shorttitle} 
\newcommand\vldbavailabilityurl{http://vldb.org/pvldb/format_vol14.html}
\newcommand\vldbpagestyle{empty}

\begin{document}




\title{RPT: Relational Pre-trained Transformer Is Almost All You Need towards Democratizing Data Preparation}

\author{Nan Tang}
\affiliation{%
  \institution{QCRI, HBKU, Qatar}
}
\email{ntang@hbku.edu.qa}

\author{Ju Fan}
\affiliation{
  \institution{Renmin University, China}
}
\email{fanj@ruc.edu.cn}

\author{Fangyi Li}
\affiliation{%
  \institution{Renmin University, China}
}
\email{fangyili@ruc.edu.cn}

\author{Jianhong Tu}
\affiliation{%
  \institution{Renmin University, China}
}
\email{tujh@ruc.edu.cn}

\author{Xiaoyong Du}
\affiliation{%
  \institution{Renmin University, China}
}
\email{duyong@ruc.edu.cn}

\author{Guoliang Li}
\affiliation{%
  \institution{Tsinghua University, China}
}
\email{liguoliang@tsinghua.edu.cn}

\author{Sam Madden}
\affiliation{%
  \institution{CSAIL, MIT, USA}
}
\email{madden@csail.mit.edu}

\author{Mourad Ouzzani}
\affiliation{%
  \institution{QCRI, HBKU, Qatar}
}
\email{mouzzani@hbku.edu.qa}


\begin{abstract}
{\em Can AI help automate human-easy but computer-hard data preparation tasks 
that burden data scientists, practitioners, and crowd workers?} 
We answer this question by presenting \system, a denoising autoencoder for {\em tuple-to-X} models (``$X$'' could be tuple, token, label, JSON, and so on). 
\system is pre-trained for a {\em tuple-to-tuple} model by corrupting the input tuple and then learning a model to reconstruct the original tuple. 
It adopts a Transformer-based neural translation architecture that consists of a bidirectional encoder (similar to BERT) and a left-to-right autoregressive decoder (similar to GPT), leading to a generalization of both BERT and GPT. The pre-trained \system can already support several common data preparation tasks such as data cleaning, auto-completion and schema matching. Better still, \system can be fine-tuned on a wide range of data preparation tasks, such as value normalization, data transformation, data annotation, etc. 
To complement \system, 
we also discuss several appealing techniques such as collaborative training and few-shot learning for entity resolution, and few-shot learning and NLP question-answering for information extraction. 
In addition, we identify a series of research opportunities to advance the field of data preparation.
\end{abstract}

\maketitle

\pagestyle{\vldbpagestyle}
\begingroup\small\noindent\raggedright\textbf{PVLDB Reference Format:}\\
\vldbauthors. \vldbtitle. PVLDB, \vldbvolume(\vldbissue): \vldbpages, \vldbyear.\\
\href{https://doi.org/\vldbdoi}{doi:\vldbdoi}
\endgroup
\begingroup
\renewcommand\thefootnote{}\footnote{\noindent
$^*$ Ju Fan is the corresponding author.\\ 

	\noindent This work is licensed under the Creative Commons BY-NC-ND 4.0 International License. Visit \url{https://creativecommons.org/licenses/by-nc-nd/4.0/} to view a copy of this license. For any use beyond those covered by this license, obtain permission by emailing \href{mailto:info@vldb.org}{info@vldb.org}. Copyright is held by the owner/author(s). Publication rights licensed to the VLDB Endowment. \\
	\raggedright Proceedings of the VLDB Endowment, Vol. \vldbvolume, No. \vldbissue\ %
	ISSN 2150-8097. \\
	\href{https://doi.org/\vldbdoi}{doi:\vldbdoi} \\
}\addtocounter{footnote}{-1}\endgroup

\ifdefempty{\vldbavailabilityurl}{}{
	\vspace{.3cm}
	\begingroup\small\noindent\raggedright\textbf{PVLDB Artifact Availability:}\\
	The source code, data, and/or other artifacts have been made available at \url{\vldbavailabilityurl}.
	\endgroup
}


\section{Introduction}
\label{sec:intro}

Data preparation --- including data cleaning~\cite{DBLP:journals/pvldb/AbedjanCDFIOPST16}, data transformation~\cite{DBLP:journals/sigmod/HameedN20}, entity resolution~\cite{DBLP:journals/tkde/ElmagarmidIV07}, information extraction~\cite{DBLP:journals/sigmod/CafarellaMH08}, and so forth --- is the most time-consuming and least enjoyable work for data  scientists~\cite{DBLP:conf/cidr/DengFAWSEIMO017}. Next, we present several scenarios to better understand these problems.

\begin{figure}[t!]
\centering
\includegraphics[width=\columnwidth]{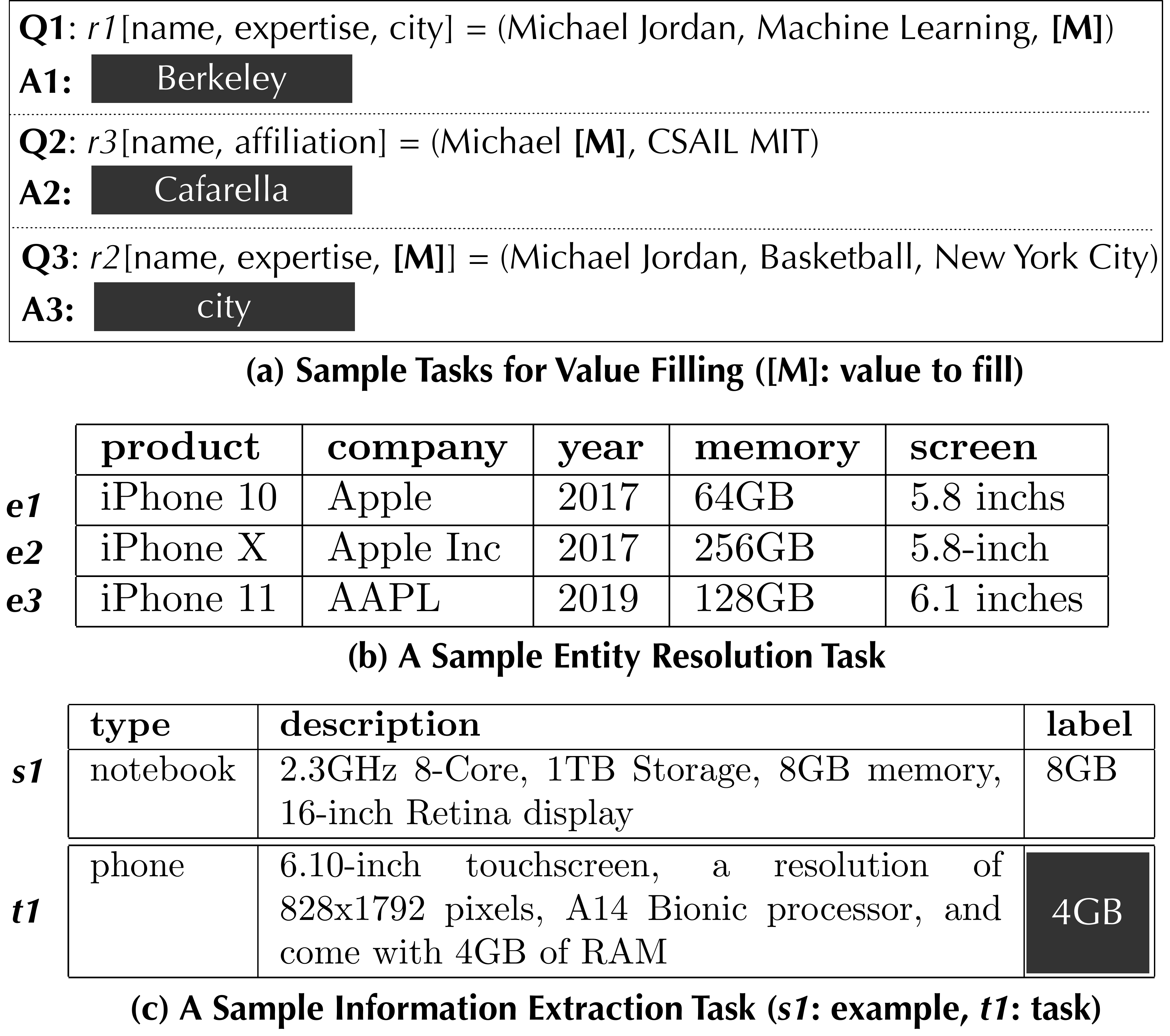}
\vspace{-2.5em}
\caption{Motivating Scenarios.}
\label{scenarios}
\vspace*{-1.5em}
\end{figure}

\vspace{0.8ex}
{\em Scenario 1: Data Cleaning.}
Figure~\ref{scenarios}(a) Q1 and Q2 show two typical data cleaning problems.
(i) {\em Cell Filling:} Question Q1 asks for the \at{city} for the ``Michael Jordan'' whose expertise is ``Machine Learning''.
(ii) {\em Value Filling:} Q2 asks for the last name of someone who works at ``CSAIL MIT'' with the first name ``Michael''.

Answering Q1 can help solve a series of problems such as error detection, data repairing, and missing value imputation; and
answering Q2 can help auto-completion (\eg give the answer A2 ``Cafarella'') and auto-suggestion  (\eg provide a list of candidate names such as \{Cafarella, Stonebraker\}).

\vspace{0.8ex}
{\em Scenario 2: Attribute Filling for Schema Matching.}
Figure~\ref{scenarios}(a) Q3 asks for the attribute name for the value ``New York City'', \wrt \at{name} ``Michael Jordan'' and \at{expertise} ``Basketball''. Answering this question can help schema matching, a core data integration problem~\cite{DBLP:books/daglib/0029346}, by better aligning attributes from different tables.

\vspace{0.8ex}
{\em Scenario 3: Entity Resolution (ER)} 
Figure~\ref{scenarios}(b) shows a typical ER task that asks whether $e_1$, $e_2$ and $e_3$ are 
the ``same''.

A human with enough knowledge can tell that ``iPhone 10'' $=$ ``iPhone X'' $\neq$ ``iPhone 11'', ``Apple'' $=$ ``Apple Inc'' $=$ ``AAPL'', and ``inches'' $=$ ``-inch''.
Hence, one can decide that $e_1$ and $e_2$ do not match $e_3$, and $e_1$ matches $e_2$ (if the memory does not matter).

\vspace{0.8ex}
{\em Scenario 4: Information Extraction (IE)}
Figure~\ref{scenarios}(c) shows an IE task, which is typically done via crowdsourcing~\cite{DBLP:journals/pvldb/JainSPW17}. A requester provides several samples (\eg $s_1$) that show what ``label'' should be extracted, and asks  workers to perform similar tasks (\eg $t_1$).

A crowd worker needs first to interpret the task by analyzing $s_1$ (and maybe a few more examples) and concretizes it as ``{\em what is the memory size}''. Afterwards, he can perform $t_1$ by extracting the label 4GB from $t_1[\at{description]}$ by knowing ``RAM'' is for memory. 

\stitle{Challenges.}
Scenarios (1--4) are simple for humans, but are hard for computers. To solve them, computers face the following challenges.
(1) {\em Knowledge:} computers need to have the background knowledge through  {\em understanding} an enormous corpora of tables.
(2) {\em Experience:} computers should be able to learn from prior and various tasks.
(3) {\em Adaptation:} computers should (quickly) adjust to new inputs and new tasks.

\stitle{Vision.}
Indeed, these problems have been seen as `holy grail' problems for the database community for decades~\cite{DBLP:journals/pvldb/AbedjanCDFIOPST16,DBLP:books/acm/IlyasC19,DBLP:series/synthesis/2012Fan,DBLP:conf/pods/GolshanHMT17,DBLP:books/daglib/0029346,DBLP:journals/pvldb/LiuCCJR10}, but despite thousands of papers on these topics, they still remain unsolved.  
Recent evidence from the NLP community, where DL-based models and representations have been shown to perform nearly as well as humans on various language understanding and question answering tasks, suggests that a learned approach may be a viable option for these 
data preparation tasks as well.

\stitle{Desiderata.}
The desiderata for AI-powered tools to achieve near-human intelligence for data preparation is summarized as below, in response to Challenges (1--3), respectively. 
(1) {\em Deep learning architecture and self-supervised pre-training.} 
	We need a deep learning architecture that can learn from many tables, similar to language models that can learn from a large text corpora. This simulates how humans gain knowledge.
	Moreover, it should be pre-trained without human provided labels, \ie self-supervision.
(2) {\em Transfer learning.} 
	The model should be able obtain knowledge from different tasks on different datasets.
(3) {\em Fine-tuning and few-shot learning.} 
	The pre-trained model should allow customization on different downstream applications through fine-tuning. 
	In addition, it should be able to understand a new task from a few examples.

\stitle{RPT Is *Almost* All You Need.} 
The design of \textit{\textbf{R}}elational \textit{\textbf{P}}re-trained \textit{\textbf{T}}ransformer (\system) is inspired by recent successes of DL models in NLP. The fundamental questions for relational data understanding on data preparation are:
(1) {\em what is the architecture?} and (2) {\em what is the surrogate task for pre-training?}

\sstab
{\em (1) \system Architecture.}
Typical choices are 
encoder-only such as BERT~\cite{bert}, 
decoder-only such as GPT-3~\cite{gpt3}, or 
encoder-decoder such as BART~\cite{bart} and T5~\cite{t5}. In fact, the encoder-decoder architecture can be considered as a generalization of the encoder-only model (\eg BERT) and the decoder-only model (\eg GPT-3). Recent studies from BART and T5 found that encoder-decoder models generally outperform encoder-only or decoder-only language models.
Thus, a Transformer-based~\cite{DBLP:conf/nips/VaswaniSPUJGKP17} encoder-decoder model provides more flexibility and can be adapted to a wide range of data preparation tasks, and hence can be used by \system.

\sstab
{\em (2) \system Pre-training.} 
There have been several works on pre-training using tables, such as TAPAS~\cite{tapas}, TURL~\cite{turl}, TaBERT~\cite{tabert} and TabFact~\cite{tabfact}.
However, since most data preparation tasks are in the granularity of tuples, instead of entire tables, we posit that training \system tuple-by-tuple is more desirable. 
For the pre-training objectives, most recent studies confirm that {\em fill-in-the-blank} style denoising objectives (where the model is trained to recover missing pieces in the input) work best; examples include 
BERT~\cite{bert}, BART~\cite{bart}, and T5~\cite{t5} for NLP, and 
TURL~\cite{turl} for relational tables.

\stitle{Contributions.} We make the following notable contributions.

\bi
	\item {\em \system}: 
	We describe a standard Transformer-based denoising autoencoder architecture to pre-train sequence-to-sequence models for {\em tuple-to-tuple} training, with new {\em tuple-aware masking} mechanisms. (Section~\ref{sec:rpt-model})

	\item {\em Fine-tuning \system:} 
	We discuss a wide range of data preparation tasks that can be supported by fine-tuning \system. (Section~\ref{sec:finetune})

	\item {\em Beyond \system:} 
	We discuss several appealing techniques that can complement \system in specific data preparation tasks, \eg collaborative training and few-shot learning for ER, and few-shot learning and NLP question-answering for IE. (Section~\ref{sec:beyond})
\ei


\section{RPT}
\label{sec:rpt-model}


\subsection{Architecture}
\label{subsec:rpt-architecture}

\system uses a standard sequence-to-sequence (or encoder-decoder) Transformer~\cite{DBLP:conf/nips/VaswaniSPUJGKP17} model, similar to BART~\cite{DBLP:conf/acl/PressSL20}, as shown in Figure~\ref{rpt}.

{\em Encoder.} \system uses a bidirectional encoder (similar to BERT~\cite{bert}) because it has the advantage of learning to predict the corrupted data bidirectionally, from both the context on the left and the context on the right of the corrupted data. Moreover, it is Transformer-based, which can use self-attention to generate a richer representation of each input token. Hence, a Transformer-based bidirectional encoder is a natural fit for reading tuples where, by definition, the ordering of (attribute name, attribute value) pairs is irrelevant.

{\em Decoder.}
\system uses a left-to-right autoregressive decoder (similar to GPT-3~\cite{gpt3}). 

\subsection{Pre-training \system}
\label{subsec:rpt-pretrain}

\system is pre-trained on tuples, for which we just need to corrupt tuples and then optimize a reconstruction loss -- the cross-entropy between the model output and the original tuple.


\stitle{Tuple Tokenization.} We represent each tuple as a concatenation of its attribute names and  values. For example, tuple $t_1$ in Figure~\ref{scenarios}(a) can be tokenized as: 

\vspace{1ex}
\hspace{-3ex}
\fbox{name Michael Jordan expertise Machine Learning city Berkeley}

\stitle{Token Embeddings.}
Because there is a clear semantic difference between attribute names and values, we can add special tokens, \AT~before an attribute name and \VT~before an attribute value. 
Token embeddings are widely used in NLP tasks, such as the [CLS] (indicating the start) and [SEP] (indicating the next sentence) tokens used by BERT~\cite{bert}.
Hence, we can get a sequence of $t_1$ with a richer tuple-aware semantics as:

\begin{figure*}[t!]
\centering
\includegraphics[width=0.9\textwidth]{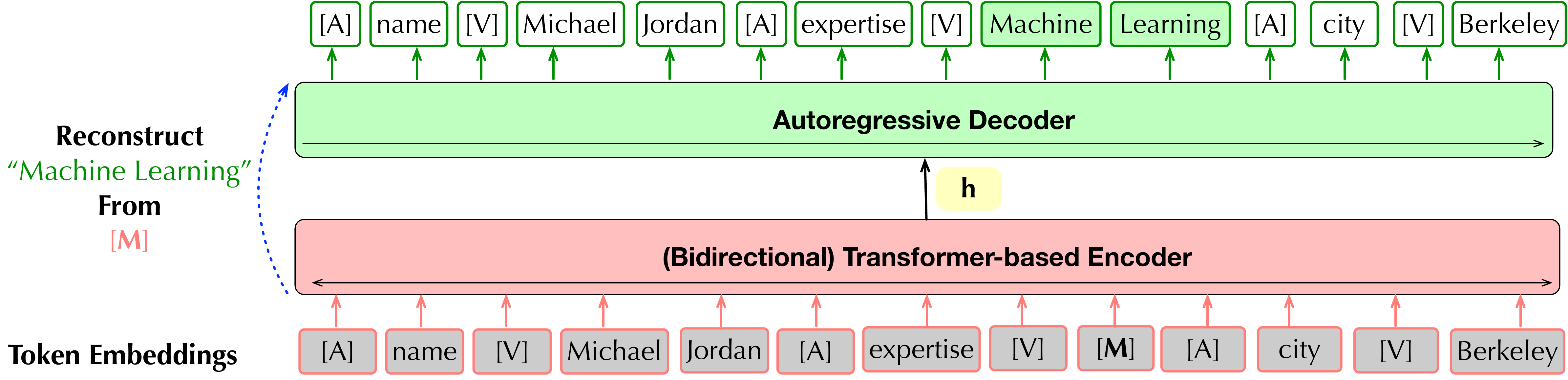}
\vspace{-1.2em}
\caption{The RPT Architecture.}
\label{rpt}
\vspace{-1em}
\end{figure*}

\vspace{1ex}
\hspace{-3ex}
\fbox{\begin{minipage}{0.98\columnwidth}
\AT~name \VT~Michael Jordan \AT~expertise \VT~Machine Learning \AT~city \VT~Berkeley
\end{minipage}
}

\stitle{Positional and Column Embeddings.}
We can add additional meta-data such as positional embeddings (\ie indicating the token's position in the sequence) 
and segment embeddings (\eg adding the same segment embedding to the tokens belonging to the same attribute value), 
which are inspired by TAPAS~\cite{tapas} for table parsing.

\stitle{Working Mechanism.}
Given an input sequence of a tuple with some value to be masked out (\eg ``Machine Learning'' in Figure~\ref{rpt}) 
represented by a ``mask token" \M, along with rich semantic information (\eg an attribute \AT~or a value \VT, which position, and which column), 
the bidirectional encoder will look at the information before and after the masked token \M, learn which token to pay attention to (using Transformer~\cite{DBLP:conf/nips/VaswaniSPUJGKP17}), and generate an intermediate vector representation $h$. The autoregressive decoder will take $h$ as input, and generate an  output sequence, by denoising the masked input \M~to be ``Machine Learning''. By doing so, we can train \system in an unsupervised fashion, without any human labels.

One difficulty is to predict how many tokens are masked by one \M.
Note that, BERT~\cite{bert} masks each token with one \M; i.e., ``Machine Learning'' will be masked as \M\M, which tells explicitly how many tokens are missing. 
We cannot do the same (masking each token with one \M), because during prediction, we do not know how many tokens are missing. 
The ability to teach the model to predict the length of the missing tokens masked by one single mask token \M~can be achieved by {\em text infilling}~\cite{DBLP:journals/tacl/JoshiCLWZL20}, which will be discussed next.

\stitle{Token Masking.} 
(1) {\em Attribute Name Masking:} We randomly select attribute names to mask, \eg 
  \at{name}.
  
(2) {\em Entire Attribute Value Masking:} We randomly select entire attribute values to mask, \eg ``Machine Learning'' is masked with one \M~(see Figure~\ref{rpt}), which forces \system to first predict the number of masked tokens and then predict the tokens.
  
(3)  {\em Single Attribute Value Masking:} We randomly select a single attribute value (\ie one token) to mask, \eg ``Jordan''.

Note that one possible optimization to the above process is as follows.
Instead of giving \system the full freedom to learn how input tokens attend on each other in the form of an attention matrix~\cite{DBLP:conf/nips/VaswaniSPUJGKP17}, 
we  add some explicit rules. 
For example, 
(i) an attribute name (\eg \at{name}) can only attend on the other attribute names (\eg \at{expertise} and \at{city}) and its associated tokens for attribute values (\eg ``Michael'' and ``Jordan''), but not other attribute values (\eg ``Berkeley''), 
and 
(ii) a token for an attribute value (\eg ``Berkeley'') can only attend to all attribute values of all attributes
and its attribute name (\ie \at{city}), but not other attribute names (\eg \at{name}). TURL~\cite{turl} also uses this technique, called {\em visibility matrix}.

\subsection{Related Work}
\label{subsec:rpt-related}


\sstab{\bf Data Cleaning.} 
We categorize prior work on data cleaning into three categories.
(i) {\em Only examine the data at hand.} 
There are integrity constraints (FDs~\cite{DBLP:conf/sigmod/BohannonFFR05}, its extensions CFDs~\cite{DBLP:journals/tods/FanGJK08} and PFDs~\cite{DBLP:journals/pvldb/QahtanTOCS20}, denial constraints~\cite{DBLP:journals/pvldb/ChuIP13}, and rule-based methods~\cite{DBLP:conf/sigmod/WangT14,DBLP:conf/sigmod/HeVSLMPT16}), and probabilistic based methods (\eg HoloClean~\cite{DBLP:journals/pvldb/RekatsinasCIR17}). They need enough signals or data redundancy from $D$.
Supervised ML based methods (\eg GDR~\cite{DBLP:journals/pvldb/YakoutENOI11}, SCAREd~\cite{DBLP:conf/sigmod/YakoutBE13}, Raha~\cite{DBLP:conf/sigmod/MahdaviAFMOS019} and Baran~\cite{DBLP:journals/pvldb/MahdaviA20}) learn only from the data at hand, which cannot be generalized to other datasets.
(ii) {\em Use External Reliable Sources:} This includes the of master data~\cite{DBLP:journals/pvldb/FanLMTY10,DBLP:conf/icde/InterlandiT15} or knowledge bases~\cite{DBLP:conf/sigmod/ChuMIOP0Y15,DBLP:conf/icde/Hao0LL17}.
These methods require experts to define or confirm the matching between the data at hand and the external source, \eg matching rules for table-table matching~\cite{DBLP:journals/pvldb/FanLMTY10} or graphical patterns for table-graph matching~\cite{DBLP:conf/sigmod/ChuMIOP0Y15}.
(iii) {\em Human- or crowd-in-the-loop.}
When neither (i) nor (ii) type solutions work, a last resort is to fall back on humans to 
clean the dataset.

Intuitively, with enough signals, data redundancy, reliable external sources with sufficient coverage, and the availability of experts to bootstrap and tune the process, 
we can leverage  (i) and (ii) style solutions. 
Unfortunately, this is usually not the case in practice~\cite{DBLP:journals/pvldb/AbedjanCDFIOPST16} ---  cleaning  is frequently of type (iii) with its high human cost. Automating type (iii) solutions is the main motivation of \system.

\stitle{Knowledge Bases.} 
There are two ways to encode the knowledge: 
``explicit knowledge'' such as knowledge graphs, 
or ``implicit knowledge'' by {\em memorizing} the knowledge using DL models. 
In practice, both explicit knowledge graphs and implicit pre-trained DL models have been widely studied in industry and academia. 
Both directions are important, and \system belongs to the latter. 
One drawback is that it is hard to explain, for which explainable AI techniques~\cite{DBLP:journals/corr/Doshi-VelezK17,imltutorial,DBLP:conf/kdd/Ribeiro0G16} will play an important role for grounding \system-like tools.

\stitle{Relational Table Understanding.} TaBERT~\cite{tabert}, Tapas~\cite{tapas} and TabFact~\cite{tabfact} study the question-answering tasks that involve joint reasoning over both free text and structured tables. They take a natural language utterance as input and produce a structured query (\eg an SQL query in TaBERT or aggregations in Tapas~\cite{tapas}), or a classification result (\eg support or refute in TabFact). To this end, they focus on learning a joint representation over textual utterances and tabular data with Transformer models and designing various pre-training tasks to this end. %

Closer to this work is TURL~\cite{turl}. However, \system differs from TURL in two aspects:
(i) TURL employs an encoder-only architecture for learned representations, instead of generating a complicated output, \eg a tuple or a table. The additional decoder architecture of \system provides the flexibility of generating sequences in multiple forms.
(ii) TURL has to be used with a KB to extract values. For example, for cell infilling, TURL uses a pre-trained model (1.2GB) to generate a representation, which has to be linked to the KB (\ie a collection of web tables, 4.6GB) to get the actual value. \system (1.6GB in our experiment) does not need such a KB to fill missing values.

\subsection{Opportunities}
\label{subsec:rpt-opp}

\system naturally supports several common data preparation tasks, \eg error detection, data repairing, auto-completion, auto-suggestion, and schema matching.
Yet there are also many opportunities.

\stab
(O1) {\em Hybrid Solutions.}
While \system does not differentiate between categorical or numeric data during pre-training, it works better for categorical data (\ie human-easy). 
A promising  direction is to combine \system with other (quantitative) data cleaning methods~\cite{DBLP:journals/pvldb/ProkoshynaSCMS15} from a rich set of  (a-b) type data cleaning solutions.

\stab
(O2) {\em Dirty Data.}
Many tables are dirty. Pre-training \system on these dirty tables may yield a biased result. 
Currently, we learn directly from dirty tables, by assuming that the frequency of correct values is higher than the frequency of wrong values. 
There are several open problems.
First, we would like to provide some guarantee of model robustness while still learning from dirty data.
Second,  a cleaned version of training data that can be used as a benchmark is highly desired, similar to the Colossal Clean Crawled Corpus (C4) for Text-To-Text-Transfer-Transformer (T5)~\cite{t5}.

\stab
(O3) {\em An AI-assisted Tool with Human-in-the-loop.}
Achieving high accuracy in diverse data preparation tasks and domains is still a challenge for \system; it would require substantial  in-domain training data. Hence, a practical usage of \system is to use it as an AI-assisted tool that can suggest meaningful results in many human-in-the-loop tasks, which can guide users and thus reduce human cost.

\subsection{Preliminary Result}
\label{subsec:rpt-exp}

We have conducted preliminary experiments to show that \system can reconstruct the masked token(s) in tuples. Our baseline is BART~\cite{DBLP:conf/acl/PressSL20}, which is pre-trained with a large corpus of text, including from the product domain. Because BART and \system have the same architecture (Figure~\ref{rpt}), we can use the parameters pre-trained by BART, instead of a random initialization. We used tables about products, including Abt-Buy~\cite{abtbuy} and Walmart-Amazon~\cite{walmartamazon}.  
Note that these two tables are naturally dirty.

\begin{figure*}[t!]
\centering
\includegraphics[width=0.8\textwidth]{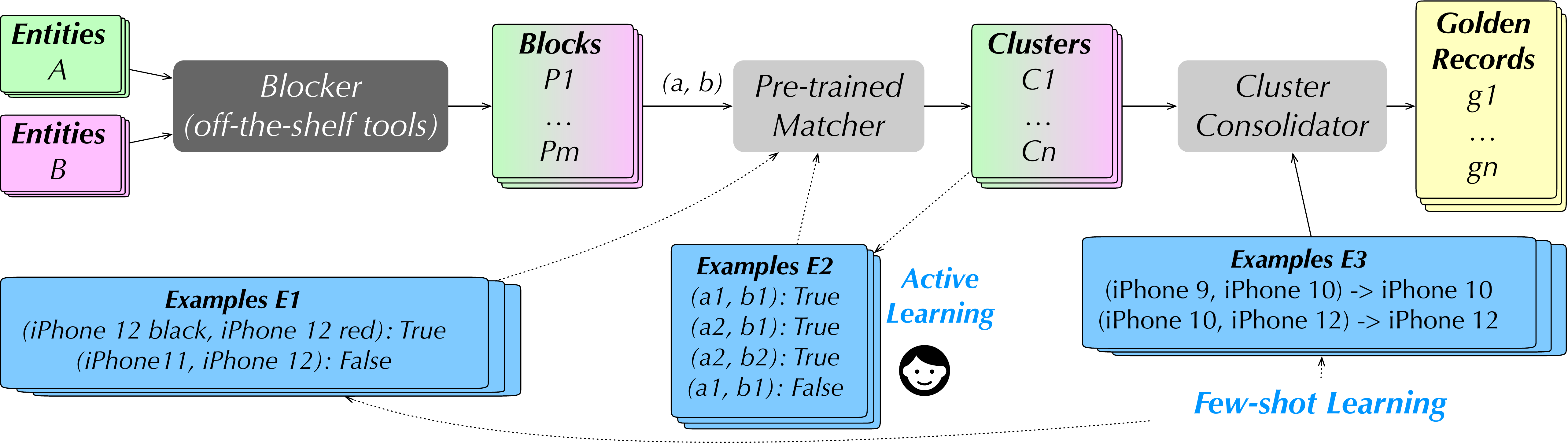}
\vspace{-1.0em}
\caption{Collaborative Learning and Few-shot Learning for Entity Resolution.}
\label{rpt-e}
\vspace{-1em}
\end{figure*}

\begin{table}[t!]
\caption{Compare \system with BART (yellow: masked values; green: (partially) correct; pink: wrong).}
\label{tbl:rptc-expt}
\vspace{-1em}
\resizebox{\columnwidth}{!}{
\begin{tabular}{|p{20mm}|p{19mm}|l|p{13mm}|p{11mm}|l|}
  \hline
  {\bf title} & {\bf manufacturer} & {\bf price} & \cellcolor{green!25}{\bf Truth} & {\bf \system-C} & {\bf BART} \\
  \hline
  instant home design (jewel case) & topics entertainment & \cellcolor{yellow!25}\M & \cellcolor{green!25}9.99 & \cellcolor{green!10}9 & \cellcolor{red!10}Topics \\
  \hline
  disney's 1st \& 2nd grade bundle ... & disney & \cellcolor{yellow!25}\M & \cellcolor{green!25}14.99 & \cellcolor{green!10}19 & \cellcolor{red!10}Dis \\
  \hline
  adobe after effects professional 6.5 ... & \cellcolor{yellow!25}\M & 499.99 & \cellcolor{green!25}adobe & \cellcolor{green!25}adobe & \cellcolor{red!10}\$1.99 \\
  \hline
  stomp inc recover lost data 2005 & \cellcolor{yellow!25}\M & 39.95 & \cellcolor{green!25}stomp inc & \cellcolor{green!10}stomp & \cellcolor{red!10}39.95 \\
  \hline 
  \cellcolor{yellow!25}\M & write brothers & 269.99 & \cellcolor{green!25}write brothers dramatica ... & \cellcolor{green!10}write brothers & \cellcolor{red!10}$1.99$ \\
  \hline
\end{tabular}
}
\vspace{-1em}
\end{table}

For testing, we used Amazon-Google~\cite{amazongoogle}, the tables that were not seen by BART or \system. We masked attribute values and asked BART and \system to predict the original values. Table~\ref{tbl:rptc-expt} shows some results, where \M~means that the value is masked out, column Truth is the ground truth, and columns \system and BART provide the results predicted by each, respectively.
Tuples 1-2 involve predicting missing prices, where \system gives close predictions but BART does not.
Tuples 3-4 involve predicting  missing manufacturers; \system-C provides good predictions.
Tuple 5 involves predicting a missing title, and \system provides a partially correct prediction.

This preliminary experiment shows that \system pre-trained  on tables can learn  structural data values from tables better than directly using a pre-trained language model (\eg BART), which is not customized for relational data.
The main reason is that, by pre-training on the product tables, \system can better learn dependency among columns, and thus is more capable of predicting missing values.
Of course, \system sometimes makes wrong predictions, but for those cases, BART  also fails.  
Our belief is that these preliminary results are suggestive enough of the effectiveness of the approach that it merits significant additional investigation.

\stitle{Limitations.} \system  faces similar limitations that pre-trained language models (LMs) face. 
(1) {\em Numeric values:}  numeric values are usually mapped into unknown tokens causing the model to fail on tasks that require precise prediction on numeric values.
(2) {\em Max sequence length:} restricted by the GPU memory size, most pre-trained LMs are limited by the sequence length, thus data preparation on wide tables may require additional optimization.
(3) {\em Not fully reliable.} Similar to GPT-3, a generative model cannot be fully trusted. One way to combat this is to treat it as an AI-assistant with a human-in-the-loop, as discussed in Section~\ref{subsec:rpt-opp} Opportunities (O3).

\section{Fine-tuning RPT}
\label{sec:finetune}

The encoder-decoder architecture of \system (pre-trained on {\em tuple-to-tuple}) provides the flexibility to be fine-tuned for different downstream data preparation tasks (\ie~{\em tuple-to-X}).

\stitle{Value Normalization.}
Because \system has an autoregressive decoder, it can be directly fine-tuned for sequence generation tasks such as value normalization (\eg ``Mike Jordan, 9 ST, Berkeley'' $\ra$ ``Mike Jordan, 9th Street, Berkeley''). The encoder takes the input value as a sequence and the decoder generates the output autoregressively.
In addition, normalizing ``Mike'' to ``Michael'' or ``Sam'' to ``Samuel'' can be fine-tuned as a neural name translation~\cite{DBLP:conf/coling/UgawaTNTO18} task.

\stitle{Data Transformation.}
Similarly to what is described above, \system can be fine-tuned for transformation of data from one format (\eg a tuple) to another format (\eg JSON or XML), where the decoder will autoregressively serialize the output in the target format.

\stitle{Data Annotation.}
Given a tuple, data annotation requires adding a label (\eg a classification task). We can use the final hidden state of the final decoder token to fine-tune a multi-class linear classifier.

\noindent
{\bf Information Extraction (IE).}
Given a tuple, IE extracts a span or multiple spans of relevant text, which can be done by fine-tuning the decoder to produce the (start, end) pairs of spans.

\stitle{Learned Tuple Representation for Entity Resolution.}
The embeddings of entities have been used in entity resolution for both blocking~\cite{DBLP:journals/pvldb/EbraheemTJOT18} and entity matching~\cite{DBLP:conf/sigmod/MudgalLRDPKDAR18,DBLP:journals/pvldb/EbraheemTJOT18}. A typical trick is to do cross-tuple training (or contrastive learning~\cite{DBLP:conf/icml/ChenK0H20}), via Siamese NNs~\cite{siamese}, such that similar entities have similar embeddings. Similarly, the encoder of \system can be  fine-tuned in Siamese NNs for learned representations \wrt entity resolution.

\section{Beyond RPT}
\label{sec:beyond}

In this section, we explore other, but related, techniques that can help on specific tasks.

\subsection{Entity Resolution}
\label{subsec:rpt-e}

Given two sets of entities, $A$ and $B$, an end-to-end  of {\bf\em entity resolution pipeline} (Figure~\ref{rpt-e}) is:
(1) find duplicated entity pairs $(a \in A, b \in B)$ (blocking to improve efficiency);
(2) merge them into clusters, typically through transitive closure, and 
(3) consolidate each cluster into one entity. 

\etitle{Blocking.}
There is a rich literature on automatic blocking for ER (see~\cite{DBLP:journals/csur/PapadakisSTP20} for a survey). 
There are also DL-based methods~\cite{DBLP:journals/pvldb/EbraheemTJOT18,DBLP:journals/corr/abs-2004-00584,DBLP:journals/corr/abs-2009-07203} to generate blocks. These prior works are automatic and already work well, hence will not be covered in this paper.

\etitle{Matcher.}
The state-of-the-art matchers are all ML based \eg random forests (\eg Magellan~\cite{DBLP:journals/pvldb/KondaDCDABLPZNP16}), or DL based (\eg DeepMatcher~\cite{DBLP:conf/sigmod/MudgalLRDPKDAR18} and DeepER~\cite{DBLP:journals/pvldb/EbraheemTJOT18}).
Recent works~\cite{DBLP:journals/corr/abs-2004-00584,DBLP:conf/edbt/BrunnerS20} also study to leverage pre-trained LM models for generating entity representations.

\etitle{Consolidator.}
There are rule-based methods~\cite{DBLP:conf/icde/FanGTY13} and learning-based approach~\cite{DBLP:journals/corr/abs-2006-10208} for entity consolidation -- both need either significant human involvement or a large amount of training data.

\stitle{Vision and Opportunities.}
The Matcher and Consolidator should be able to perform effectively through pre-trained models (\ie to obtain knowledge) and a few examples (\ie to interpret the task).

However, this pipeline cannot be fully automated, because some judgments are {\em objective}, \eg ``iPhone 10'', ``iPhone ten'', and ``iPhone X'' are the same, while some others are {\em subjective}, \eg whether ``iPhone 12 red'' matches ``iPhone 12 black'' is user dependent.

Our intuition is that the objective criteria can be pre-trained (such as ``iPhone 10'' matches ``iPhone X'' and ``Mike'' matches ``Michael''), but the subjective criteria need task-specific samples, for both the Matcher and the Consolidator (Figure~\ref{rpt-e}), which could be achieved by getting a few examples from humans.

We identify two major opportunities for the entire ER pipeline.

\sstab
(O1) {\em Collaborative learning} or {\em Federated Learning (FL)~\cite{DBLP:conf/mlsys/BonawitzEGHIIKK19,DBLP:journals/tist/YangLCT19}.} This is to learn the ``objective'' criteria for the Matcher. Note that there are many public and private ER benchmarks, which share common domains. It is promising to collaboratively train one Matcher, and the knowledge can be learned and transferred from one dataset to another dataset. Better still, this can be done securely~\cite{DBLP:journals/expert/LiuKXCY20}, without data sharing. Note that, there have been transfer learning techniques on ER~\cite{DBLP:journals/corr/abs-1809-11084,DBLP:conf/acl/KasaiQGLP19} to show an early success on this thread.

We believe that we should build a platform collaboratively for ER, with a pre-trained model $M$ for each domain. Anyone who wants to benefit from $M$ can download $M$, retrain using his/her data to get a $M_1$, and send back an update of parameters $\Delta_1 = M_1 - M$, and the platform will merge the model update with $M$, from multiple users~\cite{DBLP:conf/mlsys/BonawitzEGHIIKK19}. Because different entities may have different schemas, we use a pre-trained model such as BERT to be schema-agnostic.

\sstab
(O2) {\em Few-shot Learning}. 
This is to learn the ``subjective'' criteria, for Matcher and Consolidator, through a  human-in-the-loop approach.
The goal is to infer a better specified task from a few examples, \eg using Pattern-Exploiting Training~\cite{DBLP:journals/corr/abs-2001-07676}.

[Matcher.] Consider $E_1$ in Fig.~\ref{rpt-e} that contains two user provided examples and we want to automatically generate a clearer task for workers, \eg ``color does not matter but model matters''.
We can design two templates like 
(T1) ``True: if \fbox{$a$} and \fbox{b} have the same \M$_1$'' and 
(T2) ``False: if \fbox{$a$} and \fbox{b} have different \M$_2$''. 
By replacing the first matching pair in $E_1$ to template (T1), we can infer a pattern ``model'' or ``series'' (but not ``color'') for \M$_1$.
Similarly, by using the second un-matching pair in $E_2$ to template (T2), we can infer a pattern ``model'' or ``series'' for \M$_2$.

Moreover, when merging matching entities into clusters based on transitive closure, conflict may 
be automatically detected within clusters (\eg $E_2$ in Fig.~\ref{rpt-e}); such conflicts can be resolved by the users through active learning. Note that, doing active learning from conflicting predictions is different from traditional active learning methods on ER that use confusing/informative entity pairs~\cite{DBLP:journals/tkdd/BellareIPR13,DBLP:conf/pkdd/ChristenCR19}.

[Consolidator.]
Consider $E_3$ with two examples, ``iPhone 10 is more preferred than iPhone 9'', and ``iPhone 12 is more preferred than iPhone 10''. 
We can use them to make the task clearer by asking questions ``iPhone 10 is \M~than iPhone 9'' and ``iPhone 12 is \M~than iPhone 10'', and enforce a language model to fill the two masked tokens with the same value, which might be ``newer''.

Another powerful method related to few-shot learning is meta-learning (or learning to learn fast)~\cite{DBLP:books/sp/98/ThrunP98}, with the main goal to learn new concepts and skills fast with a few  examples.

\stitle{Preliminary Results.}
We have conducted some preliminary experiments on the ``product'' domain for the Matcher, because this domain has a rich collection of ER benchmarks, and it is known to be hard ER cases because they are text-rich. Specifically, we use five well-known benchmarks. (D1) Abt-Buy~\cite{abtbuy}, (D2) Amazon-Google~\cite{amazongoogle}, (D3) Walmart-Amazon~\cite{walmartamazon}, (D4) iTunes-Amazon~\cite{itunesamazon}, and (D5) SIGMOD 2020 programming contest~\cite{sigmod20} (we took 1000/8000 matching/unmatching pairs). 


Specifically, when testing on D1, we train with D2--D5, and when testing on D2, we train with D1, D3--D5. 
We only tested on D1 and D2, so we can directly compare with ZeroER~\cite{DBLP:conf/sigmod/WuCSCT20} (without any training data) and DeepMatcher~\cite{DBLP:conf/sigmod/MudgalLRDPKDAR18} (trained with hundreds/thousands of examples), where the F1 scores and the number of labels are reported from their original papers.
We also compare with Ditto~\cite{DBLP:journals/corr/abs-2004-00584}. We progressively add the number of labeled data to fine-tune Ditto until its F1 is very close to that of collaborative training (CT).

We also note that the paper~\cite{DBLP:conf/edbt/BrunnerS20} is quite similar to Ditto. First, both studies use the same strategy to model a record pair as a sequence and fine-tune a pre-trained model to output the matching results. Second, \cite{DBLP:conf/edbt/BrunnerS20} compares different pre-trained LM models for entity resolution and reports that RoBERTa~\cite{DBLP:journals/pvldb/EbraheemTJOT18} achieves the best performance. Similarly, the latest version of Ditto~\cite{DBLP:journals/corr/abs-2004-00584} also uses RoBERTa as its pre-trained model. Third, both studies achieve similar results in the experiments. For example, on the Abt-Buy dataset, \cite{DBLP:conf/edbt/BrunnerS20} and Ditto achieve $0.91$ and $0.90$ on F1 score respectively.


Table~\ref{tbl:rpte-expt} shows that CT outperforms ZeroER and is comparable with DeepMatcher that was trained with 1000+ examples. Moreover, CT uses zero examples from the test ER dataset to achieve the performance of Ditto, which is trained by fine-tuning a pre-trained model with 1000+ examples.
This result  verifies the opportunity (O1)  that it is promising to collaboratively train a Matcher to decide whether two entities (even in different schemas) match or not.

\begin{table}[t!]
\caption{Comparison with the State of the art.}
\label{tbl:rpte-expt}
\vspace{-1em}
	\resizebox{\columnwidth}{!}{%
	\begin{tabular}{|l||c|c|c|c|}
	\hline
	\multirow{2}{*}{\textbf{Method}} &
	\multicolumn{2}{c|}{\textbf{Abt-Buy}} &
	\multicolumn{2}{c|}{\textbf{Amazon-Google}} 
	\\
	\cline{2-5}
	& ~~\textbf{F1 score}~~ &~~\textbf{\# labels}~~ &~~\textbf{F1 score}~~ & ~~\# \textbf{labels}~~  \\  \hline	
	\hline
	{\bf Collaborative Training (CT)} & 0.72 & 0 & 0.53 & 0 \\ \hline 
	{\bf ZeroER} & 0.52 & 0 & 0.48 & 0 \\ \hline 
	{\bf DeepMatcher} & 0.63 & 7689 & 0.69 & 9167 \\ \hline
	{\bf {Ditto}} & 0.71 & 1500 & 0.50 & 1000 \\ \hline
\end{tabular}
}
\vspace{-1em}
\end{table}

\subsection{Information Extraction}
\label{sec:rpt-i}

Information Extraction (IE) is the process of retrieving specific information from unstructured (\eg text), or (semi-)structured (\eg relational) data. We consider a simple IE problem: given a text or text-rich tuple $t$, it is to extract a span (\ie a continuous sequence of tokens) from $t$, denoted by ${\bf I}(t)$.
See Figure~\ref{scenarios}(c) for a sample IE task.
Although simple, the above definition is general enough to cover a wide range of IE problems.

\sstab
{\bf Connection with Question-Answering in NLP.}
There is a natural connection between the IE problem and a typical question-answering problem in NLP. For question answering, there is an input question such as ``Q: where do water droplets collide with ice crystals to form precipitation'', and an input paragraph ``P: ... Precipitation forms as smaller droplets coalesce via collision with other rain drop or ice crystals within a cloud. ...''. The task is to find a span \eg \fbox{within a cloud} of paragraph P to answer the question Q. There are many NLP question-answering benchmarks, such as SQuAD~\cite{DBLP:conf/emnlp/RajpurkarZLL16} where AI has outperformed human performance~\cite{zhang2020retrospective}.

\begin{figure}[t!]
\vspace{-1em}
\centering
\includegraphics[width=0.85\columnwidth]{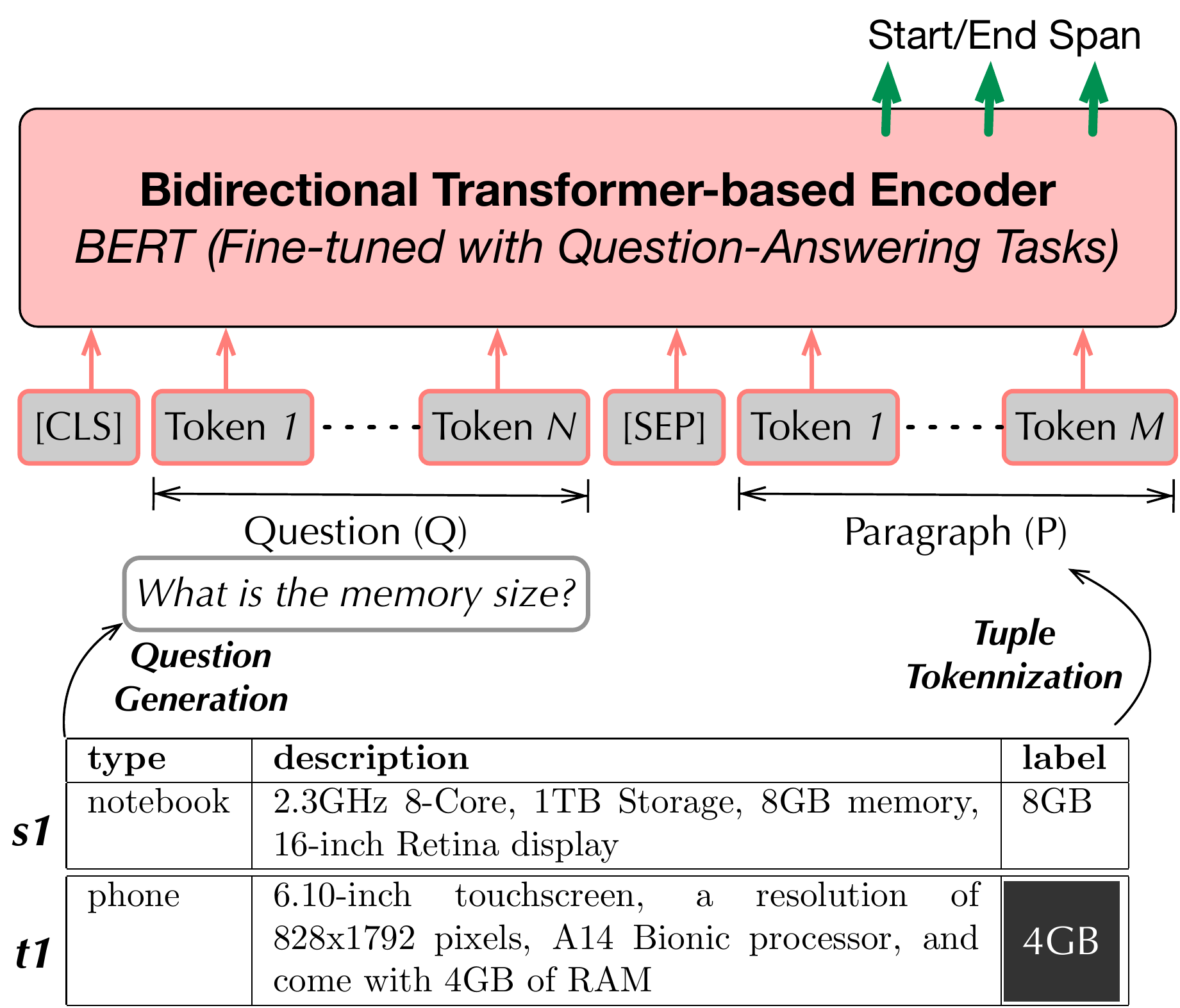}
\vspace{-1.0em}
\caption{Connecting IE with NLP Question-Answering.}
\label{rpt-i}
\vspace{-2.5em}
\end{figure}

%
As shown in Figure~\ref{rpt-i}, given a query Q and a paragraph P, a pre-trained model fine-tuned using question-answering benchmarks can provide a span, \ie  (start, end) positions, as output. 

We can tokenize a tuple $t$ as a paragraph $P$ (Section~\ref{sec:rpt-model}).
The remaining problem is to generate the question Q.
We can have a question template such as ``what is the \M'', where the \M~can be instantiated with one-shot learning (\eg the label of $s_1$) via \eg PET~\cite{DBLP:journals/corr/abs-2001-07676}, which gives ``what is the memory size'' as the question Q.


\etitle{Opportunities.} 
(O1) Connect more DB-related IE tasks to well-studied NLP tasks, so as to obtain pre-trained knowledge (\eg NeruON~\cite{DBLP:conf/naacl/BhutaniSTHJ19} uses a seq-to-seq model to extract tuples from question-answer pairs).
(O2) Currently, many IE tasks are performed by crowd workers (or crowd-in-the-loop). Instead of fully replacing these crowd workers, we are studying how to train multiple \system-I models as AI-workers, and mix the AI-workers and crowd workers to reduce the total cost of a crowdsourcing task.

\section{Call to Arms}
\label{sec:conclusion}

We have presented our vision and concrete steps for democratizing data preparation: 
\system, fine-tuning \system, and other appealing techniques. Several recent successes (\eg Termite~\cite{DBLP:conf/sigmod/FernandezM19}, \textsc{EmbDI}~\cite{DBLP:conf/sigmod/CappuzzoPT20}, TURL~\cite{turl}, Ditto~\cite{DBLP:journals/corr/abs-2004-00584} and NeurON~\cite{DBLP:conf/naacl/BhutaniSTHJ19}) have shed some light on this direction. Our preliminary results, along with these related papers suggest that learning-based approaches have the potential to outperform more traditional methods, much as they have revolutionized NLP.
However, the data preparation field is vast, the problems are diverse and much work remains to be done.
In particular, a major obstacle to advance all the above topics 
is the limited availability of real-world benchmarks, \eg C4 for T5~\cite{t5}. Now is the time for the data preparation and larger database communities to come together to explore the potential of these new techniques. 

\begin{acks}
 This work was partly supported by National Key Research and Development Program of China (2020YFB2104101), NSF of China (61632016, 61925205, 62072461, U1911203), Huawei, TAL Education, and Beijing National Research Center for Information Science and Technology.
\end{acks}

\bibliographystyle{ACM-Reference-Format}
\bibliography{DA}


\begin{thebibliography}{73}


\ifx \showCODEN    \undefined \def \showCODEN     #1{\unskip}     \fi
\ifx \showDOI      \undefined \def \showDOI       #1{#1}\fi
\ifx \showISBNx    \undefined \def \showISBNx     #1{\unskip}     \fi
\ifx \showISBNxiii \undefined \def \showISBNxiii  #1{\unskip}     \fi
\ifx \showISSN     \undefined \def \showISSN      #1{\unskip}     \fi
\ifx \showLCCN     \undefined \def \showLCCN      #1{\unskip}     \fi
\ifx \shownote     \undefined \def \shownote      #1{#1}          \fi
\ifx \showarticletitle \undefined \def \showarticletitle #1{#1}   \fi
\ifx \showURL      \undefined \def \showURL       {\relax}        \fi
\providecommand\bibfield[2]{#2}
\providecommand\bibinfo[2]{#2}
\providecommand\natexlab[1]{#1}
\providecommand\showeprint[2][]{arXiv:#2}

\bibitem[\protect\citeauthoryear{Abedjan, Chu, Deng, Fernandez, Ilyas, Ouzzani,
  Papotti, Stonebraker, and Tang}{Abedjan et~al\mbox{.}}{2016}]%
        {DBLP:journals/pvldb/AbedjanCDFIOPST16}
\bibfield{author}{\bibinfo{person}{Ziawasch Abedjan}, \bibinfo{person}{Xu Chu},
  \bibinfo{person}{Dong Deng}, \bibinfo{person}{Raul~Castro Fernandez},
  \bibinfo{person}{Ihab~F. Ilyas}, \bibinfo{person}{Mourad Ouzzani},
  \bibinfo{person}{Paolo Papotti}, \bibinfo{person}{Michael Stonebraker}, {and}
  \bibinfo{person}{Nan Tang}.} \bibinfo{year}{2016}\natexlab{}.
\newblock \showarticletitle{Detecting Data Errors: Where are we and what needs
  to be done?}
\newblock \bibinfo{journal}{\emph{Proc. {VLDB} Endow.}} \bibinfo{volume}{9},
  \bibinfo{number}{12} (\bibinfo{year}{2016}), \bibinfo{pages}{993--1004}.
\newblock


\bibitem[\protect\citeauthoryear{{Abt-Buy}}{{Abt-Buy}}{[n.d.]}]%
        {abtbuy}
\bibfield{author}{\bibinfo{person}{{Abt-Buy}}.}
  \bibinfo{year}{[n.d.]}\natexlab{}.
\newblock
\newblock
\newblock
\shownote{\url{https://github.com/anhaidgroup/deepmatcher/blob/master/Datasets.md\#abt-buy}.}


\bibitem[\protect\citeauthoryear{{Amazon-Google}}{{Amazon-Google}}{[n.d.]}]%
        {amazongoogle}
\bibfield{author}{\bibinfo{person}{{Amazon-Google}}.}
  \bibinfo{year}{[n.d.]}\natexlab{}.
\newblock
\newblock
\newblock
\shownote{\url{https://github.com/anhaidgroup/deepmatcher/blob/master/Datasets.md\#amazon-google}.}


\bibitem[\protect\citeauthoryear{Bellare, Iyengar, Parameswaran, and
  Rastogi}{Bellare et~al\mbox{.}}{2013}]%
        {DBLP:journals/tkdd/BellareIPR13}
\bibfield{author}{\bibinfo{person}{Kedar Bellare}, \bibinfo{person}{Suresh
  Iyengar}, \bibinfo{person}{Aditya~G. Parameswaran}, {and}
  \bibinfo{person}{Vibhor Rastogi}.} \bibinfo{year}{2013}\natexlab{}.
\newblock \showarticletitle{Active Sampling for Entity Matching with
  Guarantees}.
\newblock \bibinfo{journal}{\emph{{ACM} Trans. Knowl. Discov. Data}}
  \bibinfo{volume}{7}, \bibinfo{number}{3} (\bibinfo{year}{2013}),
  \bibinfo{pages}{12:1--12:24}.
\newblock


\bibitem[\protect\citeauthoryear{Bhutani, Suhara, Tan, Halevy, and
  Jagadish}{Bhutani et~al\mbox{.}}{2019}]%
        {DBLP:conf/naacl/BhutaniSTHJ19}
\bibfield{author}{\bibinfo{person}{Nikita Bhutani}, \bibinfo{person}{Yoshihiko
  Suhara}, \bibinfo{person}{Wang{-}Chiew Tan}, \bibinfo{person}{Alon~Y.
  Halevy}, {and} \bibinfo{person}{H.~V. Jagadish}.}
  \bibinfo{year}{2019}\natexlab{}.
\newblock \showarticletitle{Open Information Extraction from Question-Answer
  Pairs}. In \bibinfo{booktitle}{\emph{{NAACL-HLT}}},
  \bibfield{editor}{\bibinfo{person}{Jill Burstein}, \bibinfo{person}{Christy
  Doran}, {and} \bibinfo{person}{Thamar Solorio}} (Eds.).
  \bibinfo{pages}{2294--2305}.
\newblock


\bibitem[\protect\citeauthoryear{Bohannon, Flaster, Fan, and Rastogi}{Bohannon
  et~al\mbox{.}}{2005}]%
        {DBLP:conf/sigmod/BohannonFFR05}
\bibfield{author}{\bibinfo{person}{Philip Bohannon}, \bibinfo{person}{Michael
  Flaster}, \bibinfo{person}{Wenfei Fan}, {and} \bibinfo{person}{Rajeev
  Rastogi}.} \bibinfo{year}{2005}\natexlab{}.
\newblock \showarticletitle{A Cost-Based Model and Effective Heuristic for
  Repairing Constraints by Value Modification}. In
  \bibinfo{booktitle}{\emph{SIGMOD}}, \bibfield{editor}{\bibinfo{person}{Fatma
  {\"{O}}zcan}} (Ed.). \bibinfo{pages}{143--154}.
\newblock


\bibitem[\protect\citeauthoryear{Bonawitz, Eichner, Grieskamp, Huba, Ingerman,
  Ivanov, Kiddon, Konecn{\'{y}}, Mazzocchi, McMahan, Overveldt, Petrou, Ramage,
  and Roselander}{Bonawitz et~al\mbox{.}}{2019}]%
        {DBLP:conf/mlsys/BonawitzEGHIIKK19}
\bibfield{author}{\bibinfo{person}{Keith Bonawitz}, \bibinfo{person}{Hubert
  Eichner}, \bibinfo{person}{Wolfgang Grieskamp}, \bibinfo{person}{Dzmitry
  Huba}, \bibinfo{person}{Alex Ingerman}, \bibinfo{person}{Vladimir Ivanov},
  \bibinfo{person}{Chlo{\'{e}} Kiddon}, \bibinfo{person}{Jakub Konecn{\'{y}}},
  \bibinfo{person}{Stefano Mazzocchi}, \bibinfo{person}{Brendan McMahan},
  \bibinfo{person}{Timon~Van Overveldt}, \bibinfo{person}{David Petrou},
  \bibinfo{person}{Daniel Ramage}, {and} \bibinfo{person}{Jason Roselander}.}
  \bibinfo{year}{2019}\natexlab{}.
\newblock \showarticletitle{Towards Federated Learning at Scale: System
  Design}. In \bibinfo{booktitle}{\emph{{MLSys}}}.
\newblock


\bibitem[\protect\citeauthoryear{Brown, Mann, Ryder, Subbiah, Kaplan, Dhariwal,
  Neelakantan, Shyam, Sastry, Askell, Agarwal, Herbert{-}Voss, Krueger,
  Henighan, Child, Ramesh, Ziegler, Wu, Winter, Hesse, Chen, Sigler, Litwin,
  Gray, Chess, Clark, Berner, McCandlish, Radford, Sutskever, and Amodei}{Brown
  et~al\mbox{.}}{2020}]%
        {gpt3}
\bibfield{author}{\bibinfo{person}{Tom~B. Brown}, \bibinfo{person}{Benjamin
  Mann}, \bibinfo{person}{Nick Ryder}, \bibinfo{person}{Melanie Subbiah},
  \bibinfo{person}{Jared Kaplan}, \bibinfo{person}{Prafulla Dhariwal},
  \bibinfo{person}{Arvind Neelakantan}, \bibinfo{person}{Pranav Shyam},
  \bibinfo{person}{Girish Sastry}, \bibinfo{person}{Amanda Askell},
  \bibinfo{person}{Sandhini Agarwal}, \bibinfo{person}{Ariel Herbert{-}Voss},
  \bibinfo{person}{Gretchen Krueger}, \bibinfo{person}{Tom Henighan},
  \bibinfo{person}{Rewon Child}, \bibinfo{person}{Aditya Ramesh},
  \bibinfo{person}{Daniel~M. Ziegler}, \bibinfo{person}{Jeffrey Wu},
  \bibinfo{person}{Clemens Winter}, \bibinfo{person}{Christopher Hesse},
  \bibinfo{person}{Mark Chen}, \bibinfo{person}{Eric Sigler},
  \bibinfo{person}{Mateusz Litwin}, \bibinfo{person}{Scott Gray},
  \bibinfo{person}{Benjamin Chess}, \bibinfo{person}{Jack Clark},
  \bibinfo{person}{Christopher Berner}, \bibinfo{person}{Sam McCandlish},
  \bibinfo{person}{Alec Radford}, \bibinfo{person}{Ilya Sutskever}, {and}
  \bibinfo{person}{Dario Amodei}.} \bibinfo{year}{2020}\natexlab{}.
\newblock \showarticletitle{Language Models are Few-Shot Learners}. In
  \bibinfo{booktitle}{\emph{{NeurIPS}}}.
\newblock


\bibitem[\protect\citeauthoryear{Brunner and Stockinger}{Brunner and
  Stockinger}{2020}]%
        {DBLP:conf/edbt/BrunnerS20}
\bibfield{author}{\bibinfo{person}{Ursin Brunner} {and} \bibinfo{person}{Kurt
  Stockinger}.} \bibinfo{year}{2020}\natexlab{}.
\newblock \showarticletitle{Entity Matching with Transformer Architectures -
  {A} Step Forward in Data Integration}. In \bibinfo{booktitle}{\emph{{EDBT}}}.
  \bibinfo{pages}{463--473}.
\newblock


\bibitem[\protect\citeauthoryear{Cafarella, Madhavan, and Halevy}{Cafarella
  et~al\mbox{.}}{2008}]%
        {DBLP:journals/sigmod/CafarellaMH08}
\bibfield{author}{\bibinfo{person}{Michael~J. Cafarella},
  \bibinfo{person}{Jayant Madhavan}, {and} \bibinfo{person}{Alon~Y. Halevy}.}
  \bibinfo{year}{2008}\natexlab{}.
\newblock \showarticletitle{Web-scale extraction of structured data}.
\newblock \bibinfo{journal}{\emph{{SIGMOD} Rec.}} \bibinfo{volume}{37},
  \bibinfo{number}{4} (\bibinfo{year}{2008}), \bibinfo{pages}{55--61}.
\newblock


\bibitem[\protect\citeauthoryear{Cappuzzo, Papotti, and
  Thirumuruganathan}{Cappuzzo et~al\mbox{.}}{2020}]%
        {DBLP:conf/sigmod/CappuzzoPT20}
\bibfield{author}{\bibinfo{person}{Riccardo Cappuzzo}, \bibinfo{person}{Paolo
  Papotti}, {and} \bibinfo{person}{Saravanan Thirumuruganathan}.}
  \bibinfo{year}{2020}\natexlab{}.
\newblock \showarticletitle{Creating Embeddings of Heterogeneous Relational
  Datasets for Data Integration Tasks}. In
  \bibinfo{booktitle}{\emph{{SIGMOD}}}. \bibinfo{pages}{1335--1349}.
\newblock


\bibitem[\protect\citeauthoryear{Chen, Kornblith, Norouzi, and Hinton}{Chen
  et~al\mbox{.}}{2020}]%
        {DBLP:conf/icml/ChenK0H20}
\bibfield{author}{\bibinfo{person}{Ting Chen}, \bibinfo{person}{Simon
  Kornblith}, \bibinfo{person}{Mohammad Norouzi}, {and}
  \bibinfo{person}{Geoffrey~E. Hinton}.} \bibinfo{year}{2020}\natexlab{}.
\newblock \showarticletitle{A Simple Framework for Contrastive Learning of
  Visual Representations}. In \bibinfo{booktitle}{\emph{{ICML}}}.
\newblock


\bibitem[\protect\citeauthoryear{Chen, Wang, Chen, Zhang, Wang, Li, Zhou, and
  Wang}{Chen et~al\mbox{.}}{2019}]%
        {tabfact}
\bibfield{author}{\bibinfo{person}{Wenhu Chen}, \bibinfo{person}{Hongmin Wang},
  \bibinfo{person}{Jianshu Chen}, \bibinfo{person}{Yunkai Zhang},
  \bibinfo{person}{Hong Wang}, \bibinfo{person}{Shiyang Li},
  \bibinfo{person}{Xiyou Zhou}, {and} \bibinfo{person}{William~Yang Wang}.}
  \bibinfo{year}{2019}\natexlab{}.
\newblock \showarticletitle{TabFact: {A} Large-scale Dataset for Table-based
  Fact Verification}.
\newblock \bibinfo{journal}{\emph{CoRR}}  \bibinfo{volume}{abs/1909.02164}
  (\bibinfo{year}{2019}).
\newblock


\bibitem[\protect\citeauthoryear{Chicco}{Chicco}{2021}]%
        {siamese}
\bibfield{author}{\bibinfo{person}{Davide Chicco}.}
  \bibinfo{year}{2021}\natexlab{}.
\newblock \showarticletitle{Siamese Neural Networks: An Overview}.
\newblock In \bibinfo{booktitle}{\emph{Artificial Neural Networks - Third
  Edition}}. \bibinfo{series}{Methods in Molecular Biology},
  Vol.~\bibinfo{volume}{2190}. \bibinfo{pages}{73--94}.
\newblock


\bibitem[\protect\citeauthoryear{Christen, Christen, and Rahm}{Christen
  et~al\mbox{.}}{2019}]%
        {DBLP:conf/pkdd/ChristenCR19}
\bibfield{author}{\bibinfo{person}{Victor Christen}, \bibinfo{person}{Peter
  Christen}, {and} \bibinfo{person}{Erhard Rahm}.}
  \bibinfo{year}{2019}\natexlab{}.
\newblock \showarticletitle{Informativeness-Based Active Learning for Entity
  Resolution}. In \bibinfo{booktitle}{\emph{{ECML} {PKDD}}},
  \bibfield{editor}{\bibinfo{person}{Peggy Cellier} {and} \bibinfo{person}{Kurt
  Driessens}} (Eds.), Vol.~\bibinfo{volume}{1168}. \bibinfo{pages}{125--141}.
\newblock


\bibitem[\protect\citeauthoryear{Chu, Ilyas, and Papotti}{Chu
  et~al\mbox{.}}{2013}]%
        {DBLP:journals/pvldb/ChuIP13}
\bibfield{author}{\bibinfo{person}{Xu Chu}, \bibinfo{person}{Ihab~F. Ilyas},
  {and} \bibinfo{person}{Paolo Papotti}.} \bibinfo{year}{2013}\natexlab{}.
\newblock \showarticletitle{Discovering Denial Constraints}.
\newblock \bibinfo{journal}{\emph{Proc. {VLDB} Endow.}} \bibinfo{volume}{6},
  \bibinfo{number}{13} (\bibinfo{year}{2013}), \bibinfo{pages}{1498--1509}.
\newblock


\bibitem[\protect\citeauthoryear{Chu, Morcos, Ilyas, Ouzzani, Papotti, Tang,
  and Ye}{Chu et~al\mbox{.}}{2015}]%
        {DBLP:conf/sigmod/ChuMIOP0Y15}
\bibfield{author}{\bibinfo{person}{Xu Chu}, \bibinfo{person}{John Morcos},
  \bibinfo{person}{Ihab~F. Ilyas}, \bibinfo{person}{Mourad Ouzzani},
  \bibinfo{person}{Paolo Papotti}, \bibinfo{person}{Nan Tang}, {and}
  \bibinfo{person}{Yin Ye}.} \bibinfo{year}{2015}\natexlab{}.
\newblock \showarticletitle{{KATARA:} {A} Data Cleaning System Powered by
  Knowledge Bases and Crowdsourcing}. In \bibinfo{booktitle}{\emph{{SIGMOD}}}.
  \bibinfo{pages}{1247--1261}.
\newblock


\bibitem[\protect\citeauthoryear{Deng, Fernandez, Abedjan, Wang, Stonebraker,
  Elmagarmid, Ilyas, Madden, Ouzzani, and Tang}{Deng et~al\mbox{.}}{2017}]%
        {DBLP:conf/cidr/DengFAWSEIMO017}
\bibfield{author}{\bibinfo{person}{Dong Deng}, \bibinfo{person}{Raul~Castro
  Fernandez}, \bibinfo{person}{Ziawasch Abedjan}, \bibinfo{person}{Sibo Wang},
  \bibinfo{person}{Michael Stonebraker}, \bibinfo{person}{Ahmed~K. Elmagarmid},
  \bibinfo{person}{Ihab~F. Ilyas}, \bibinfo{person}{Samuel Madden},
  \bibinfo{person}{Mourad Ouzzani}, {and} \bibinfo{person}{Nan Tang}.}
  \bibinfo{year}{2017}\natexlab{}.
\newblock \showarticletitle{The Data Civilizer System}. In
  \bibinfo{booktitle}{\emph{CIDR}}.
\newblock


\bibitem[\protect\citeauthoryear{Deng, Sun, Lees, Wu, and Yu}{Deng
  et~al\mbox{.}}{2020}]%
        {turl}
\bibfield{author}{\bibinfo{person}{Xiang Deng}, \bibinfo{person}{Huan Sun},
  \bibinfo{person}{Alyssa Lees}, \bibinfo{person}{You Wu}, {and}
  \bibinfo{person}{Cong Yu}.} \bibinfo{year}{2020}\natexlab{}.
\newblock \showarticletitle{{TURL:} Table Understanding through Representation
  Learning}.
\newblock \bibinfo{journal}{\emph{Proc. {VLDB} Endow.}} (\bibinfo{year}{2020}).
\newblock


\bibitem[\protect\citeauthoryear{Devlin, Chang, Lee, and Toutanova}{Devlin
  et~al\mbox{.}}{[n.d.]}]%
        {bert}
\bibfield{author}{\bibinfo{person}{Jacob Devlin}, \bibinfo{person}{Ming{-}Wei
  Chang}, \bibinfo{person}{Kenton Lee}, {and} \bibinfo{person}{Kristina
  Toutanova}.} \bibinfo{year}{[n.d.]}\natexlab{}.
\newblock \showarticletitle{{BERT:} Pre-training of Deep Bidirectional
  Transformers for Language Understanding}. In
  \bibinfo{booktitle}{\emph{NAACL-HLT}},
  \bibfield{editor}{\bibinfo{person}{Jill Burstein}, \bibinfo{person}{Christy
  Doran}, {and} \bibinfo{person}{Thamar Solorio}} (Eds.).
  \bibinfo{pages}{4171--4186}.
\newblock


\bibitem[\protect\citeauthoryear{Doan, Halevy, and Ives}{Doan
  et~al\mbox{.}}{2012}]%
        {DBLP:books/daglib/0029346}
\bibfield{author}{\bibinfo{person}{AnHai Doan}, \bibinfo{person}{Alon~Y.
  Halevy}, {and} \bibinfo{person}{Zachary~G. Ives}.}
  \bibinfo{year}{2012}\natexlab{}.
\newblock \bibinfo{booktitle}{\emph{Principles of Data Integration}}.
\newblock \bibinfo{publisher}{Morgan Kaufmann}.
\newblock


\bibitem[\protect\citeauthoryear{Doshi{-}Velez and Kim}{Doshi{-}Velez and
  Kim}{2017}]%
        {DBLP:journals/corr/Doshi-VelezK17}
\bibfield{author}{\bibinfo{person}{Finale Doshi{-}Velez} {and}
  \bibinfo{person}{Been Kim}.} \bibinfo{year}{2017}\natexlab{}.
\newblock \showarticletitle{A Roadmap for a Rigorous Science of
  Interpretability}.
\newblock \bibinfo{journal}{\emph{CoRR}}  \bibinfo{volume}{abs/1702.08608}
  (\bibinfo{year}{2017}).
\newblock


\bibitem[\protect\citeauthoryear{Ebraheem, Thirumuruganathan, Joty, Ouzzani,
  and Tang}{Ebraheem et~al\mbox{.}}{2018}]%
        {DBLP:journals/pvldb/EbraheemTJOT18}
\bibfield{author}{\bibinfo{person}{Muhammad Ebraheem},
  \bibinfo{person}{Saravanan Thirumuruganathan}, \bibinfo{person}{Shafiq~R.
  Joty}, \bibinfo{person}{Mourad Ouzzani}, {and} \bibinfo{person}{Nan Tang}.}
  \bibinfo{year}{2018}\natexlab{}.
\newblock \showarticletitle{Distributed Representations of Tuples for Entity
  Resolution}.
\newblock \bibinfo{journal}{\emph{Proc. {VLDB} Endow.}} \bibinfo{volume}{11},
  \bibinfo{number}{11} (\bibinfo{year}{2018}), \bibinfo{pages}{1454--1467}.
\newblock


\bibitem[\protect\citeauthoryear{Elmagarmid, Ipeirotis, and
  Verykios}{Elmagarmid et~al\mbox{.}}{2007}]%
        {DBLP:journals/tkde/ElmagarmidIV07}
\bibfield{author}{\bibinfo{person}{Ahmed~K. Elmagarmid},
  \bibinfo{person}{Panagiotis~G. Ipeirotis}, {and}
  \bibinfo{person}{Vassilios~S. Verykios}.} \bibinfo{year}{2007}\natexlab{}.
\newblock \showarticletitle{Duplicate Record Detection: {A} Survey}.
\newblock \bibinfo{journal}{\emph{{IEEE} Trans. Knowl. Data Eng.}}
  \bibinfo{volume}{19}, \bibinfo{number}{1} (\bibinfo{year}{2007}),
  \bibinfo{pages}{1--16}.
\newblock


\bibitem[\protect\citeauthoryear{Fan and Geerts}{Fan and Geerts}{2012}]%
        {DBLP:series/synthesis/2012Fan}
\bibfield{author}{\bibinfo{person}{Wenfei Fan} {and} \bibinfo{person}{Floris
  Geerts}.} \bibinfo{year}{2012}\natexlab{}.
\newblock \bibinfo{booktitle}{\emph{Foundations of Data Quality Management}}.
\newblock \bibinfo{publisher}{Morgan {\&} Claypool Publishers}.
\newblock


\bibitem[\protect\citeauthoryear{Fan, Geerts, Jia, and Kementsietsidis}{Fan
  et~al\mbox{.}}{2008}]%
        {DBLP:journals/tods/FanGJK08}
\bibfield{author}{\bibinfo{person}{Wenfei Fan}, \bibinfo{person}{Floris
  Geerts}, \bibinfo{person}{Xibei Jia}, {and} \bibinfo{person}{Anastasios
  Kementsietsidis}.} \bibinfo{year}{2008}\natexlab{}.
\newblock \showarticletitle{Conditional functional dependencies for capturing
  data inconsistencies}.
\newblock \bibinfo{journal}{\emph{{ACM} Trans. Database Syst.}}
  \bibinfo{volume}{33}, \bibinfo{number}{2} (\bibinfo{year}{2008}),
  \bibinfo{pages}{6:1--6:48}.
\newblock


\bibitem[\protect\citeauthoryear{Fan, Geerts, Tang, and Yu}{Fan
  et~al\mbox{.}}{2013}]%
        {DBLP:conf/icde/FanGTY13}
\bibfield{author}{\bibinfo{person}{Wenfei Fan}, \bibinfo{person}{Floris
  Geerts}, \bibinfo{person}{Nan Tang}, {and} \bibinfo{person}{Wenyuan Yu}.}
  \bibinfo{year}{2013}\natexlab{}.
\newblock \showarticletitle{Inferring data currency and consistency for
  conflict resolution}. In \bibinfo{booktitle}{\emph{{ICDE}}},
  \bibfield{editor}{\bibinfo{person}{Christian~S. Jensen},
  \bibinfo{person}{Christopher~M. Jermaine}, {and} \bibinfo{person}{Xiaofang
  Zhou}} (Eds.). \bibinfo{pages}{470--481}.
\newblock


\bibitem[\protect\citeauthoryear{Fan, Li, Ma, Tang, and Yu}{Fan
  et~al\mbox{.}}{2010}]%
        {DBLP:journals/pvldb/FanLMTY10}
\bibfield{author}{\bibinfo{person}{Wenfei Fan}, \bibinfo{person}{Jianzhong Li},
  \bibinfo{person}{Shuai Ma}, \bibinfo{person}{Nan Tang}, {and}
  \bibinfo{person}{Wenyuan Yu}.} \bibinfo{year}{2010}\natexlab{}.
\newblock \showarticletitle{Towards Certain Fixes with Editing Rules and Master
  Data}.
\newblock \bibinfo{journal}{\emph{Proc. {VLDB} Endow.}} \bibinfo{volume}{3},
  \bibinfo{number}{1} (\bibinfo{year}{2010}), \bibinfo{pages}{173--184}.
\newblock


\bibitem[\protect\citeauthoryear{Fernandez and Madden}{Fernandez and
  Madden}{2019}]%
        {DBLP:conf/sigmod/FernandezM19}
\bibfield{author}{\bibinfo{person}{Raul~Castro Fernandez} {and}
  \bibinfo{person}{Samuel Madden}.} \bibinfo{year}{2019}\natexlab{}.
\newblock \showarticletitle{Termite: a system for tunneling through
  heterogeneous data}. In \bibinfo{booktitle}{\emph{Proceedings of the Second
  International Workshop on Exploiting Artificial Intelligence Techniques for
  Data Management, aiDM@SIGMOD 2019, Amsterdam, The Netherlands, July 5,
  2019}}. \bibinfo{pages}{7:1--7:8}.
\newblock


\bibitem[\protect\citeauthoryear{Golshan, Halevy, Mihaila, and Tan}{Golshan
  et~al\mbox{.}}{2017}]%
        {DBLP:conf/pods/GolshanHMT17}
\bibfield{author}{\bibinfo{person}{Behzad Golshan}, \bibinfo{person}{Alon~Y.
  Halevy}, \bibinfo{person}{George~A. Mihaila}, {and}
  \bibinfo{person}{Wang{-}Chiew Tan}.} \bibinfo{year}{2017}\natexlab{}.
\newblock \showarticletitle{Data Integration: After the Teenage Years}. In
  \bibinfo{booktitle}{\emph{{PODS}}},
  \bibfield{editor}{\bibinfo{person}{Emanuel Sallinger},
  \bibinfo{person}{Jan~Van den Bussche}, {and} \bibinfo{person}{Floris Geerts}}
  (Eds.). \bibinfo{publisher}{{ACM}}, \bibinfo{pages}{101--106}.
\newblock


\bibitem[\protect\citeauthoryear{Hameed and Naumann}{Hameed and
  Naumann}{2020}]%
        {DBLP:journals/sigmod/HameedN20}
\bibfield{author}{\bibinfo{person}{Mazhar Hameed} {and} \bibinfo{person}{Felix
  Naumann}.} \bibinfo{year}{2020}\natexlab{}.
\newblock \showarticletitle{Data Preparation: {A} Survey of Commercial Tools}.
\newblock \bibinfo{journal}{\emph{{SIGMOD} Rec.}} \bibinfo{volume}{49},
  \bibinfo{number}{3} (\bibinfo{year}{2020}), \bibinfo{pages}{18--29}.
\newblock


\bibitem[\protect\citeauthoryear{Hao, Tang, Li, and Li}{Hao
  et~al\mbox{.}}{2017}]%
        {DBLP:conf/icde/Hao0LL17}
\bibfield{author}{\bibinfo{person}{Shuang Hao}, \bibinfo{person}{Nan Tang},
  \bibinfo{person}{Guoliang Li}, {and} \bibinfo{person}{Jian Li}.}
  \bibinfo{year}{2017}\natexlab{}.
\newblock \showarticletitle{Cleaning Relations Using Knowledge Bases}. In
  \bibinfo{booktitle}{\emph{{ICDE}}}. \bibinfo{pages}{933--944}.
\newblock


\bibitem[\protect\citeauthoryear{He, Veltri, Santoro, Li, Mecca, Papotti, and
  Tang}{He et~al\mbox{.}}{2016}]%
        {DBLP:conf/sigmod/HeVSLMPT16}
\bibfield{author}{\bibinfo{person}{Jian He}, \bibinfo{person}{Enzo Veltri},
  \bibinfo{person}{Donatello Santoro}, \bibinfo{person}{Guoliang Li},
  \bibinfo{person}{Giansalvatore Mecca}, \bibinfo{person}{Paolo Papotti}, {and}
  \bibinfo{person}{Nan Tang}.} \bibinfo{year}{2016}\natexlab{}.
\newblock \showarticletitle{Interactive and Deterministic Data Cleaning}. In
  \bibinfo{booktitle}{\emph{{SIGMOD}}}. \bibinfo{pages}{893--907}.
\newblock


\bibitem[\protect\citeauthoryear{Heidari, Michalopoulos, Kushagra, Ilyas, and
  Rekatsinas}{Heidari et~al\mbox{.}}{2020}]%
        {DBLP:journals/corr/abs-2006-10208}
\bibfield{author}{\bibinfo{person}{Alireza Heidari}, \bibinfo{person}{George
  Michalopoulos}, \bibinfo{person}{Shrinu Kushagra}, \bibinfo{person}{Ihab~F.
  Ilyas}, {and} \bibinfo{person}{Theodoros Rekatsinas}.}
  \bibinfo{year}{2020}\natexlab{}.
\newblock \showarticletitle{Record fusion: {A} learning approach}.
\newblock \bibinfo{journal}{\emph{CoRR}}  \bibinfo{volume}{abs/2006.10208}
  (\bibinfo{year}{2020}).
\newblock


\bibitem[\protect\citeauthoryear{Herzig, Nowak, M{\"{u}}ller, Piccinno, and
  Eisenschlos}{Herzig et~al\mbox{.}}{2020}]%
        {tapas}
\bibfield{author}{\bibinfo{person}{Jonathan Herzig},
  \bibinfo{person}{Pawel~Krzysztof Nowak}, \bibinfo{person}{Thomas
  M{\"{u}}ller}, \bibinfo{person}{Francesco Piccinno}, {and}
  \bibinfo{person}{Julian~Martin Eisenschlos}.}
  \bibinfo{year}{2020}\natexlab{}.
\newblock \showarticletitle{TaPas: Weakly Supervised Table Parsing via
  Pre-training}. In \bibinfo{booktitle}{\emph{ACL}},
  \bibfield{editor}{\bibinfo{person}{Dan Jurafsky}, \bibinfo{person}{Joyce
  Chai}, \bibinfo{person}{Natalie Schluter}, {and} \bibinfo{person}{Joel~R.
  Tetreault}} (Eds.). \bibinfo{pages}{4320--4333}.
\newblock


\bibitem[\protect\citeauthoryear{Ilyas and Chu}{Ilyas and Chu}{2019}]%
        {DBLP:books/acm/IlyasC19}
\bibfield{author}{\bibinfo{person}{Ihab~F. Ilyas} {and} \bibinfo{person}{Xu
  Chu}.} \bibinfo{year}{2019}\natexlab{}.
\newblock \bibinfo{booktitle}{\emph{Data Cleaning}}.
\newblock \bibinfo{publisher}{{ACM}}.
\newblock


\bibitem[\protect\citeauthoryear{Interlandi and Tang}{Interlandi and
  Tang}{2015}]%
        {DBLP:conf/icde/InterlandiT15}
\bibfield{author}{\bibinfo{person}{Matteo Interlandi} {and}
  \bibinfo{person}{Nan Tang}.} \bibinfo{year}{2015}\natexlab{}.
\newblock \showarticletitle{Proof positive and negative in data cleaning}. In
  \bibinfo{booktitle}{\emph{{ICDE}}}. \bibinfo{pages}{18--29}.
\newblock


\bibitem[\protect\citeauthoryear{{Itunes-Amazon}}{{Itunes-Amazon}}{[n.d.]}]%
        {itunesamazon}
\bibfield{author}{\bibinfo{person}{{Itunes-Amazon}}.}
  \bibinfo{year}{[n.d.]}\natexlab{}.
\newblock
\newblock
\newblock
\shownote{\url{https://github.com/anhaidgroup/deepmatcher/blob/master/Datasets.md\#itunes-amazon}.}


\bibitem[\protect\citeauthoryear{Jain, Sarma, Parameswaran, and Widom}{Jain
  et~al\mbox{.}}{2017}]%
        {DBLP:journals/pvldb/JainSPW17}
\bibfield{author}{\bibinfo{person}{Ayush Jain}, \bibinfo{person}{Akash~Das
  Sarma}, \bibinfo{person}{Aditya~G. Parameswaran}, {and}
  \bibinfo{person}{Jennifer Widom}.} \bibinfo{year}{2017}\natexlab{}.
\newblock \showarticletitle{Understanding Workers, Developing Effective Tasks,
  and Enhancing Marketplace Dynamics: {A} Study of a Large Crowdsourcing
  Marketplace}.
\newblock \bibinfo{journal}{\emph{Proc. {VLDB} Endow.}} \bibinfo{volume}{10},
  \bibinfo{number}{7} (\bibinfo{year}{2017}), \bibinfo{pages}{829--840}.
\newblock


\bibitem[\protect\citeauthoryear{Joshi, Chen, Liu, Weld, Zettlemoyer, and
  Levy}{Joshi et~al\mbox{.}}{2020}]%
        {DBLP:journals/tacl/JoshiCLWZL20}
\bibfield{author}{\bibinfo{person}{Mandar Joshi}, \bibinfo{person}{Danqi Chen},
  \bibinfo{person}{Yinhan Liu}, \bibinfo{person}{Daniel~S. Weld},
  \bibinfo{person}{Luke Zettlemoyer}, {and} \bibinfo{person}{Omer Levy}.}
  \bibinfo{year}{2020}\natexlab{}.
\newblock \showarticletitle{SpanBERT: Improving Pre-training by Representing
  and Predicting Spans}.
\newblock \bibinfo{journal}{\emph{Trans. Assoc. Comput. Linguistics}}
  \bibinfo{volume}{8} (\bibinfo{year}{2020}), \bibinfo{pages}{64--77}.
\newblock


\bibitem[\protect\citeauthoryear{Kasai, Qian, Gurajada, Li, and Popa}{Kasai
  et~al\mbox{.}}{2019}]%
        {DBLP:conf/acl/KasaiQGLP19}
\bibfield{author}{\bibinfo{person}{Jungo Kasai}, \bibinfo{person}{Kun Qian},
  \bibinfo{person}{Sairam Gurajada}, \bibinfo{person}{Yunyao Li}, {and}
  \bibinfo{person}{Lucian Popa}.} \bibinfo{year}{2019}\natexlab{}.
\newblock \showarticletitle{Low-resource Deep Entity Resolution with Transfer
  and Active Learning}. In \bibinfo{booktitle}{\emph{{ACL}}},
  \bibfield{editor}{\bibinfo{person}{Anna Korhonen}, \bibinfo{person}{David~R.
  Traum}, {and} \bibinfo{person}{Llu{\'{\i}}s M{\`{a}}rquez}} (Eds.).
  \bibinfo{pages}{5851--5861}.
\newblock


\bibitem[\protect\citeauthoryear{Kim and Doshi{-}Velez}{Kim and
  Doshi{-}Velez}{2017}]%
        {imltutorial}
\bibfield{author}{\bibinfo{person}{Been Kim} {and} \bibinfo{person}{Finale
  Doshi{-}Velez}.} \bibinfo{year}{2017}\natexlab{}.
\newblock \showarticletitle{Interpretable Machine Learning: The fuss, the
  concrete and the questions}. In \bibinfo{booktitle}{\emph{{ICML Tutorial}}}.
\newblock


\bibitem[\protect\citeauthoryear{Konda, Das, C., Doan, Ardalan, Ballard, Li,
  Panahi, Zhang, Naughton, Prasad, Krishnan, Deep, and Raghavendra}{Konda
  et~al\mbox{.}}{2016}]%
        {DBLP:journals/pvldb/KondaDCDABLPZNP16}
\bibfield{author}{\bibinfo{person}{Pradap Konda}, \bibinfo{person}{Sanjib Das},
  \bibinfo{person}{Paul Suganthan~G. C.}, \bibinfo{person}{AnHai Doan},
  \bibinfo{person}{Adel Ardalan}, \bibinfo{person}{Jeffrey~R. Ballard},
  \bibinfo{person}{Han Li}, \bibinfo{person}{Fatemah Panahi},
  \bibinfo{person}{Haojun Zhang}, \bibinfo{person}{Jeffrey~F. Naughton},
  \bibinfo{person}{Shishir Prasad}, \bibinfo{person}{Ganesh Krishnan},
  \bibinfo{person}{Rohit Deep}, {and} \bibinfo{person}{Vijay Raghavendra}.}
  \bibinfo{year}{2016}\natexlab{}.
\newblock \showarticletitle{Magellan: Toward Building Entity Matching
  Management Systems}.
\newblock \bibinfo{journal}{\emph{Proc. {VLDB} Endow.}} \bibinfo{volume}{9},
  \bibinfo{number}{12} (\bibinfo{year}{2016}), \bibinfo{pages}{1197--1208}.
\newblock


\bibitem[\protect\citeauthoryear{Lewis, Liu, Goyal, Ghazvininejad, Mohamed,
  Levy, Stoyanov, and Zettlemoyer}{Lewis et~al\mbox{.}}{2020}]%
        {bart}
\bibfield{author}{\bibinfo{person}{Mike Lewis}, \bibinfo{person}{Yinhan Liu},
  \bibinfo{person}{Naman Goyal}, \bibinfo{person}{Marjan Ghazvininejad},
  \bibinfo{person}{Abdelrahman Mohamed}, \bibinfo{person}{Omer Levy},
  \bibinfo{person}{Veselin Stoyanov}, {and} \bibinfo{person}{Luke
  Zettlemoyer}.} \bibinfo{year}{2020}\natexlab{}.
\newblock \showarticletitle{{BART:} Denoising Sequence-to-Sequence Pre-training
  for Natural Language Generation, Translation, and Comprehension}. In
  \bibinfo{booktitle}{\emph{{ACL}}}. \bibinfo{pages}{7871--7880}.
\newblock


\bibitem[\protect\citeauthoryear{Li, Li, Suhara, Doan, and Tan}{Li
  et~al\mbox{.}}{[n.d.]}]%
        {DBLP:journals/corr/abs-2004-00584}
\bibfield{author}{\bibinfo{person}{Yuliang Li}, \bibinfo{person}{Jinfeng Li},
  \bibinfo{person}{Yoshihiko Suhara}, \bibinfo{person}{AnHai Doan}, {and}
  \bibinfo{person}{Wang{-}Chiew Tan}.} \bibinfo{year}{[n.d.]}\natexlab{}.
\newblock \showarticletitle{Deep Entity Matching with Pre-Trained Language
  Models}.
\newblock \bibinfo{journal}{\emph{Proc. {VLDB} Endow.}}
  (\bibinfo{year}{[n.\,d.]}).
\newblock


\bibitem[\protect\citeauthoryear{Liu, Chiticariu, Chu, Jagadish, and Reiss}{Liu
  et~al\mbox{.}}{2010}]%
        {DBLP:journals/pvldb/LiuCCJR10}
\bibfield{author}{\bibinfo{person}{Bin Liu}, \bibinfo{person}{Laura
  Chiticariu}, \bibinfo{person}{Vivian Chu}, \bibinfo{person}{H.~V. Jagadish},
  {and} \bibinfo{person}{Frederick Reiss}.} \bibinfo{year}{2010}\natexlab{}.
\newblock \showarticletitle{Automatic Rule Refinement for Information
  Extraction}.
\newblock \bibinfo{journal}{\emph{Proc. {VLDB} Endow.}} \bibinfo{volume}{3},
  \bibinfo{number}{1} (\bibinfo{year}{2010}), \bibinfo{pages}{588--597}.
\newblock


\bibitem[\protect\citeauthoryear{Liu, Kang, Xing, Chen, and Yang}{Liu
  et~al\mbox{.}}{2020}]%
        {DBLP:journals/expert/LiuKXCY20}
\bibfield{author}{\bibinfo{person}{Yang Liu}, \bibinfo{person}{Yan Kang},
  \bibinfo{person}{Chaoping Xing}, \bibinfo{person}{Tianjian Chen}, {and}
  \bibinfo{person}{Qiang Yang}.} \bibinfo{year}{2020}\natexlab{}.
\newblock \showarticletitle{A Secure Federated Transfer Learning Framework}.
\newblock \bibinfo{journal}{\emph{{IEEE} Intell. Syst.}} \bibinfo{volume}{35},
  \bibinfo{number}{4} (\bibinfo{year}{2020}), \bibinfo{pages}{70--82}.
\newblock


\bibitem[\protect\citeauthoryear{Mahdavi and Abedjan}{Mahdavi and
  Abedjan}{2020}]%
        {DBLP:journals/pvldb/MahdaviA20}
\bibfield{author}{\bibinfo{person}{Mohammad Mahdavi} {and}
  \bibinfo{person}{Ziawasch Abedjan}.} \bibinfo{year}{2020}\natexlab{}.
\newblock \showarticletitle{Baran: Effective Error Correction via a Unified
  Context Representation and Transfer Learning}.
\newblock \bibinfo{journal}{\emph{Proc. {VLDB} Endow.}} \bibinfo{volume}{13},
  \bibinfo{number}{11} (\bibinfo{year}{2020}), \bibinfo{pages}{1948--1961}.
\newblock


\bibitem[\protect\citeauthoryear{Mahdavi, Abedjan, Fernandez, Madden, Ouzzani,
  Stonebraker, and Tang}{Mahdavi et~al\mbox{.}}{2019}]%
        {DBLP:conf/sigmod/MahdaviAFMOS019}
\bibfield{author}{\bibinfo{person}{Mohammad Mahdavi}, \bibinfo{person}{Ziawasch
  Abedjan}, \bibinfo{person}{Raul~Castro Fernandez}, \bibinfo{person}{Samuel
  Madden}, \bibinfo{person}{Mourad Ouzzani}, \bibinfo{person}{Michael
  Stonebraker}, {and} \bibinfo{person}{Nan Tang}.}
  \bibinfo{year}{2019}\natexlab{}.
\newblock \showarticletitle{Raha: {A} Configuration-Free Error Detection
  System}. In \bibinfo{booktitle}{\emph{{SIGMOD}}}. \bibinfo{pages}{865--882}.
\newblock


\bibitem[\protect\citeauthoryear{Mudgal, Li, Rekatsinas, Doan, Park, Krishnan,
  Deep, Arcaute, and Raghavendra}{Mudgal et~al\mbox{.}}{2018}]%
        {DBLP:conf/sigmod/MudgalLRDPKDAR18}
\bibfield{author}{\bibinfo{person}{Sidharth Mudgal}, \bibinfo{person}{Han Li},
  \bibinfo{person}{Theodoros Rekatsinas}, \bibinfo{person}{AnHai Doan},
  \bibinfo{person}{Youngchoon Park}, \bibinfo{person}{Ganesh Krishnan},
  \bibinfo{person}{Rohit Deep}, \bibinfo{person}{Esteban Arcaute}, {and}
  \bibinfo{person}{Vijay Raghavendra}.} \bibinfo{year}{2018}\natexlab{}.
\newblock \showarticletitle{Deep Learning for Entity Matching: {A} Design Space
  Exploration}. In \bibinfo{booktitle}{\emph{SIGMOD}},
  \bibfield{editor}{\bibinfo{person}{Gautam Das},
  \bibinfo{person}{Christopher~M. Jermaine}, {and} \bibinfo{person}{Philip~A.
  Bernstein}} (Eds.). \bibinfo{pages}{19--34}.
\newblock


\bibitem[\protect\citeauthoryear{Papadakis, Skoutas, Thanos, and
  Palpanas}{Papadakis et~al\mbox{.}}{2020}]%
        {DBLP:journals/csur/PapadakisSTP20}
\bibfield{author}{\bibinfo{person}{George Papadakis},
  \bibinfo{person}{Dimitrios Skoutas}, \bibinfo{person}{Emmanouil Thanos},
  {and} \bibinfo{person}{Themis Palpanas}.} \bibinfo{year}{2020}\natexlab{}.
\newblock \showarticletitle{Blocking and Filtering Techniques for Entity
  Resolution: {A} Survey}.
\newblock \bibinfo{journal}{\emph{{ACM} Comput. Surv.}} \bibinfo{volume}{53},
  \bibinfo{number}{2} (\bibinfo{year}{2020}), \bibinfo{pages}{31:1--31:42}.
\newblock


\bibitem[\protect\citeauthoryear{Press, Smith, and Levy}{Press
  et~al\mbox{.}}{[n.d.]}]%
        {DBLP:conf/acl/PressSL20}
\bibfield{author}{\bibinfo{person}{Ofir Press}, \bibinfo{person}{Noah~A.
  Smith}, {and} \bibinfo{person}{Omer Levy}.}
  \bibinfo{year}{[n.d.]}\natexlab{}.
\newblock \showarticletitle{Improving Transformer Models by Reordering their
  Sublayers}. In \bibinfo{booktitle}{\emph{ACL}},
  \bibfield{editor}{\bibinfo{person}{Dan Jurafsky}, \bibinfo{person}{Joyce
  Chai}, \bibinfo{person}{Natalie Schluter}, {and} \bibinfo{person}{Joel~R.
  Tetreault}} (Eds.). \bibinfo{pages}{2996--3005}.
\newblock


\bibitem[\protect\citeauthoryear{Prokoshyna, Szlichta, Chiang, Miller, and
  Srivastava}{Prokoshyna et~al\mbox{.}}{2015}]%
        {DBLP:journals/pvldb/ProkoshynaSCMS15}
\bibfield{author}{\bibinfo{person}{Nataliya Prokoshyna},
  \bibinfo{person}{Jaroslaw Szlichta}, \bibinfo{person}{Fei Chiang},
  \bibinfo{person}{Ren{\'{e}}e~J. Miller}, {and} \bibinfo{person}{Divesh
  Srivastava}.} \bibinfo{year}{2015}\natexlab{}.
\newblock \showarticletitle{Combining Quantitative and Logical Data Cleaning}.
\newblock \bibinfo{journal}{\emph{Proc. {VLDB} Endow.}} \bibinfo{volume}{9},
  \bibinfo{number}{4} (\bibinfo{year}{2015}), \bibinfo{pages}{300--311}.
\newblock


\bibitem[\protect\citeauthoryear{Qahtan, Tang, Ouzzani, Cao, and
  Stonebraker}{Qahtan et~al\mbox{.}}{2020}]%
        {DBLP:journals/pvldb/QahtanTOCS20}
\bibfield{author}{\bibinfo{person}{Abdulhakim~Ali Qahtan}, \bibinfo{person}{Nan
  Tang}, \bibinfo{person}{Mourad Ouzzani}, \bibinfo{person}{Yang Cao}, {and}
  \bibinfo{person}{Michael Stonebraker}.} \bibinfo{year}{2020}\natexlab{}.
\newblock \showarticletitle{Pattern Functional Dependencies for Data Cleaning}.
\newblock \bibinfo{journal}{\emph{Proc. {VLDB} Endow.}} \bibinfo{volume}{13},
  \bibinfo{number}{5} (\bibinfo{year}{2020}), \bibinfo{pages}{684--697}.
\newblock


\bibitem[\protect\citeauthoryear{Raffel, Shazeer, Roberts, Lee, Narang, Matena,
  Zhou, Li, and Liu}{Raffel et~al\mbox{.}}{2020}]%
        {t5}
\bibfield{author}{\bibinfo{person}{Colin Raffel}, \bibinfo{person}{Noam
  Shazeer}, \bibinfo{person}{Adam Roberts}, \bibinfo{person}{Katherine Lee},
  \bibinfo{person}{Sharan Narang}, \bibinfo{person}{Michael Matena},
  \bibinfo{person}{Yanqi Zhou}, \bibinfo{person}{Wei Li}, {and}
  \bibinfo{person}{Peter~J. Liu}.} \bibinfo{year}{2020}\natexlab{}.
\newblock \showarticletitle{Exploring the Limits of Transfer Learning with a
  Unified Text-to-Text Transformer}.
\newblock \bibinfo{journal}{\emph{J. Mach. Learn. Res.}}  \bibinfo{volume}{21}
  (\bibinfo{year}{2020}), \bibinfo{pages}{140:1--140:67}.
\newblock


\bibitem[\protect\citeauthoryear{Rajpurkar, Zhang, Lopyrev, and
  Liang}{Rajpurkar et~al\mbox{.}}{2016}]%
        {DBLP:conf/emnlp/RajpurkarZLL16}
\bibfield{author}{\bibinfo{person}{Pranav Rajpurkar}, \bibinfo{person}{Jian
  Zhang}, \bibinfo{person}{Konstantin Lopyrev}, {and} \bibinfo{person}{Percy
  Liang}.} \bibinfo{year}{2016}\natexlab{}.
\newblock \showarticletitle{SQuAD: 100, 000+ Questions for Machine
  Comprehension of Text}. In \bibinfo{booktitle}{\emph{{EMNLP}}},
  \bibfield{editor}{\bibinfo{person}{Jian Su}, \bibinfo{person}{Xavier
  Carreras}, {and} \bibinfo{person}{Kevin Duh}} (Eds.).
  \bibinfo{pages}{2383--2392}.
\newblock


\bibitem[\protect\citeauthoryear{Rekatsinas, Chu, Ilyas, and
  R{\'{e}}}{Rekatsinas et~al\mbox{.}}{2017}]%
        {DBLP:journals/pvldb/RekatsinasCIR17}
\bibfield{author}{\bibinfo{person}{Theodoros Rekatsinas}, \bibinfo{person}{Xu
  Chu}, \bibinfo{person}{Ihab~F. Ilyas}, {and} \bibinfo{person}{Christopher
  R{\'{e}}}.} \bibinfo{year}{2017}\natexlab{}.
\newblock \showarticletitle{HoloClean: Holistic Data Repairs with Probabilistic
  Inference}.
\newblock \bibinfo{journal}{\emph{Proc. {VLDB} Endow.}} \bibinfo{volume}{10},
  \bibinfo{number}{11} (\bibinfo{year}{2017}), \bibinfo{pages}{1190--1201}.
\newblock


\bibitem[\protect\citeauthoryear{Ribeiro, Singh, and Guestrin}{Ribeiro
  et~al\mbox{.}}{2016}]%
        {DBLP:conf/kdd/Ribeiro0G16}
\bibfield{author}{\bibinfo{person}{Marco~T{\'{u}}lio Ribeiro},
  \bibinfo{person}{Sameer Singh}, {and} \bibinfo{person}{Carlos Guestrin}.}
  \bibinfo{year}{2016}\natexlab{}.
\newblock \showarticletitle{"Why Should {I} Trust You?": Explaining the
  Predictions of Any Classifier}. In \bibinfo{booktitle}{\emph{{SIGKDD}}}.
  \bibinfo{pages}{1135--1144}.
\newblock


\bibitem[\protect\citeauthoryear{Schick and Sch{\"{u}}tze}{Schick and
  Sch{\"{u}}tze}{2020}]%
        {DBLP:journals/corr/abs-2001-07676}
\bibfield{author}{\bibinfo{person}{Timo Schick} {and} \bibinfo{person}{Hinrich
  Sch{\"{u}}tze}.} \bibinfo{year}{2020}\natexlab{}.
\newblock \showarticletitle{Exploiting Cloze Questions for Few-Shot Text
  Classification and Natural Language Inference}.
\newblock \bibinfo{journal}{\emph{CoRR}}  \bibinfo{volume}{abs/2001.07676}
  (\bibinfo{year}{2020}).
\newblock


\bibitem[\protect\citeauthoryear{{sigmod-2020-contest}}{{sigmod-2020-contest}}{[n.d.]}]%
        {sigmod20}
\bibfield{author}{\bibinfo{person}{{sigmod-2020-contest}}.}
  \bibinfo{year}{[n.d.]}\natexlab{}.
\newblock
\newblock
\newblock
\shownote{\url{http://www.inf.uniroma3.it/db/sigmod2020contest/task.html}.}


\bibitem[\protect\citeauthoryear{Thirumuruganathan, Parambath, Ouzzani, Tang,
  and Joty}{Thirumuruganathan et~al\mbox{.}}{2018}]%
        {DBLP:journals/corr/abs-1809-11084}
\bibfield{author}{\bibinfo{person}{Saravanan Thirumuruganathan},
  \bibinfo{person}{Shameem Ahamed~Puthiya Parambath}, \bibinfo{person}{Mourad
  Ouzzani}, \bibinfo{person}{Nan Tang}, {and} \bibinfo{person}{Shafiq~R.
  Joty}.} \bibinfo{year}{2018}\natexlab{}.
\newblock \showarticletitle{Reuse and Adaptation for Entity Resolution through
  Transfer Learning}.
\newblock \bibinfo{journal}{\emph{CoRR}}  \bibinfo{volume}{abs/1809.11084}
  (\bibinfo{year}{2018}).
\newblock


\bibitem[\protect\citeauthoryear{Thrun and Pratt}{Thrun and Pratt}{1998}]%
        {DBLP:books/sp/98/ThrunP98}
\bibfield{author}{\bibinfo{person}{Sebastian Thrun} {and}
  \bibinfo{person}{Lorien~Y. Pratt}.} \bibinfo{year}{1998}\natexlab{}.
\newblock \showarticletitle{Learning to Learn: Introduction and Overview}.
\newblock In \bibinfo{booktitle}{\emph{Learning to Learn}},
  \bibfield{editor}{\bibinfo{person}{Sebastian Thrun} {and}
  \bibinfo{person}{Lorien~Y. Pratt}} (Eds.). \bibinfo{publisher}{Springer},
  \bibinfo{pages}{3--17}.
\newblock


\bibitem[\protect\citeauthoryear{Ugawa, Tamura, Ninomiya, Takamura, and
  Okumura}{Ugawa et~al\mbox{.}}{2018}]%
        {DBLP:conf/coling/UgawaTNTO18}
\bibfield{author}{\bibinfo{person}{Arata Ugawa}, \bibinfo{person}{Akihiro
  Tamura}, \bibinfo{person}{Takashi Ninomiya}, \bibinfo{person}{Hiroya
  Takamura}, {and} \bibinfo{person}{Manabu Okumura}.}
  \bibinfo{year}{2018}\natexlab{}.
\newblock \showarticletitle{Neural Machine Translation Incorporating Named
  Entity}. In \bibinfo{booktitle}{\emph{{COLING}}}.
  \bibinfo{pages}{3240--3250}.
\newblock


\bibitem[\protect\citeauthoryear{Vaswani, Shazeer, Parmar, Uszkoreit, Jones,
  Gomez, Kaiser, and Polosukhin}{Vaswani et~al\mbox{.}}{2017}]%
        {DBLP:conf/nips/VaswaniSPUJGKP17}
\bibfield{author}{\bibinfo{person}{Ashish Vaswani}, \bibinfo{person}{Noam
  Shazeer}, \bibinfo{person}{Niki Parmar}, \bibinfo{person}{Jakob Uszkoreit},
  \bibinfo{person}{Llion Jones}, \bibinfo{person}{Aidan~N. Gomez},
  \bibinfo{person}{Lukasz Kaiser}, {and} \bibinfo{person}{Illia Polosukhin}.}
  \bibinfo{year}{2017}\natexlab{}.
\newblock \showarticletitle{Attention is All you Need}. In
  \bibinfo{booktitle}{\emph{NIPS}}. \bibinfo{pages}{5998--6008}.
\newblock


\bibitem[\protect\citeauthoryear{{Walmart-Amazon}}{{Walmart-Amazon}}{[n.d.]}]%
        {walmartamazon}
\bibfield{author}{\bibinfo{person}{{Walmart-Amazon}}.}
  \bibinfo{year}{[n.d.]}\natexlab{}.
\newblock
\newblock
\newblock
\shownote{\url{https://github.com/anhaidgroup/deepmatcher/blob/master/Datasets.md\#walmart-amazon}.}


\bibitem[\protect\citeauthoryear{Wang and Tang}{Wang and Tang}{2014}]%
        {DBLP:conf/sigmod/WangT14}
\bibfield{author}{\bibinfo{person}{Jiannan Wang} {and} \bibinfo{person}{Nan
  Tang}.} \bibinfo{year}{2014}\natexlab{}.
\newblock \showarticletitle{Towards dependable data repairing with fixing
  rules}. In \bibinfo{booktitle}{\emph{{SIGMOD}}}. \bibinfo{publisher}{{ACM}},
  \bibinfo{pages}{457--468}.
\newblock


\bibitem[\protect\citeauthoryear{Wang, Sisman, Wei, Dong, and Ji}{Wang
  et~al\mbox{.}}{2020}]%
        {DBLP:journals/corr/abs-2009-07203}
\bibfield{author}{\bibinfo{person}{Zhengyang Wang}, \bibinfo{person}{Bunyamin
  Sisman}, \bibinfo{person}{Hao Wei}, \bibinfo{person}{Xin~Luna Dong}, {and}
  \bibinfo{person}{Shuiwang Ji}.} \bibinfo{year}{2020}\natexlab{}.
\newblock \showarticletitle{CorDEL: {A} Contrastive Deep Learning Approach for
  Entity Linkage}.
\newblock \bibinfo{journal}{\emph{ICDM}}.
\newblock


\bibitem[\protect\citeauthoryear{Wu, Chaba, Sawlani, Chu, and
  Thirumuruganathan}{Wu et~al\mbox{.}}{2020}]%
        {DBLP:conf/sigmod/WuCSCT20}
\bibfield{author}{\bibinfo{person}{Renzhi Wu}, \bibinfo{person}{Sanya Chaba},
  \bibinfo{person}{Saurabh Sawlani}, \bibinfo{person}{Xu Chu}, {and}
  \bibinfo{person}{Saravanan Thirumuruganathan}.}
  \bibinfo{year}{2020}\natexlab{}.
\newblock \showarticletitle{ZeroER: Entity Resolution using Zero Labeled
  Examples}. In \bibinfo{booktitle}{\emph{{SIGMOD}}}.
  \bibinfo{pages}{1149--1164}.
\newblock


\bibitem[\protect\citeauthoryear{Yakout, Berti{-}{\'{E}}quille, and
  Elmagarmid}{Yakout et~al\mbox{.}}{2013}]%
        {DBLP:conf/sigmod/YakoutBE13}
\bibfield{author}{\bibinfo{person}{Mohamed Yakout}, \bibinfo{person}{Laure
  Berti{-}{\'{E}}quille}, {and} \bibinfo{person}{Ahmed~K. Elmagarmid}.}
  \bibinfo{year}{2013}\natexlab{}.
\newblock \showarticletitle{Don't be SCAREd: use SCalable Automatic REpairing
  with maximal likelihood and bounded changes}. In
  \bibinfo{booktitle}{\emph{SIGMOD}},
  \bibfield{editor}{\bibinfo{person}{Kenneth~A. Ross}, \bibinfo{person}{Divesh
  Srivastava}, {and} \bibinfo{person}{Dimitris Papadias}} (Eds.).
  \bibinfo{publisher}{{ACM}}, \bibinfo{pages}{553--564}.
\newblock


\bibitem[\protect\citeauthoryear{Yakout, Elmagarmid, Neville, Ouzzani, and
  Ilyas}{Yakout et~al\mbox{.}}{2011}]%
        {DBLP:journals/pvldb/YakoutENOI11}
\bibfield{author}{\bibinfo{person}{Mohamed Yakout}, \bibinfo{person}{Ahmed~K.
  Elmagarmid}, \bibinfo{person}{Jennifer Neville}, \bibinfo{person}{Mourad
  Ouzzani}, {and} \bibinfo{person}{Ihab~F. Ilyas}.}
  \bibinfo{year}{2011}\natexlab{}.
\newblock \showarticletitle{Guided data repair}.
\newblock \bibinfo{journal}{\emph{Proc. {VLDB} Endow.}} \bibinfo{volume}{4},
  \bibinfo{number}{5} (\bibinfo{year}{2011}), \bibinfo{pages}{279--289}.
\newblock


\bibitem[\protect\citeauthoryear{Yang, Liu, Chen, and Tong}{Yang
  et~al\mbox{.}}{2019}]%
        {DBLP:journals/tist/YangLCT19}
\bibfield{author}{\bibinfo{person}{Qiang Yang}, \bibinfo{person}{Yang Liu},
  \bibinfo{person}{Tianjian Chen}, {and} \bibinfo{person}{Yongxin Tong}.}
  \bibinfo{year}{2019}\natexlab{}.
\newblock \showarticletitle{Federated Machine Learning: Concept and
  Applications}.
\newblock \bibinfo{journal}{\emph{{ACM} Trans. Intell. Syst. Technol.}}
  \bibinfo{volume}{10}, \bibinfo{number}{2} (\bibinfo{year}{2019}),
  \bibinfo{pages}{12:1--12:19}.
\newblock


\bibitem[\protect\citeauthoryear{Yin, Neubig, Yih, and Riedel}{Yin
  et~al\mbox{.}}{2020}]%
        {tabert}
\bibfield{author}{\bibinfo{person}{Pengcheng Yin}, \bibinfo{person}{Graham
  Neubig}, \bibinfo{person}{Wen{-}tau Yih}, {and} \bibinfo{person}{Sebastian
  Riedel}.} \bibinfo{year}{2020}\natexlab{}.
\newblock \showarticletitle{TaBERT: Pretraining for Joint Understanding of
  Textual and Tabular Data}. In \bibinfo{booktitle}{\emph{{ACL}}}.
  \bibinfo{pages}{8413--8426}.
\newblock


\bibitem[\protect\citeauthoryear{Zhang, Yang, and Zhao}{Zhang
  et~al\mbox{.}}{2020}]%
        {zhang2020retrospective}
\bibfield{author}{\bibinfo{person}{Zhuosheng Zhang}, \bibinfo{person}{Junjie
  Yang}, {and} \bibinfo{person}{Hai Zhao}.} \bibinfo{year}{2020}\natexlab{}.
\newblock \bibinfo{title}{Retrospective Reader for Machine Reading
  Comprehension}.
\newblock
\newblock
\showeprint[arxiv]{2001.09694}~[cs.CL]


\end{thebibliography}

\end{document}